\documentclass[]{interact}

\usepackage[latin1]{inputenc}
\usepackage[english]{babel}

\usepackage{multirow}
\usepackage{tabularx}

\usepackage{graphicx}
\usepackage{rotating}

\usepackage{amsmath,amsfonts,amssymb,amsthm}

\usepackage{bm,bbm}
\usepackage[binary-units]{siunitx}
\usepackage{url}

\usepackage{mathrsfs}
\usepackage{color}
\usepackage{colortbl}

\usepackage[table]{xcolor}
\usepackage{lscape}
\usepackage{placeins}

\usepackage{tikz}
\usetikzlibrary{matrix}

\usepackage{adjustbox}

\makeatletter
\newcommand{\thickhline}{\noalign {\ifnum 0=`}\fi \hrule height 1pt \futurelet \reserved@a \@xhline}
\newcolumntype{"}{@{\hskip\tabcolsep\vrule width 1pt\hskip\tabcolsep}}
\makeatother

\newcommand\hphi{\mbox{$h\mbox{-}\phi$}}

\makeatletter
  \newcommand\tinyv{\@setfontsize\tinyv{5pt}{7}}
\makeatother

\usepackage{epstopdf}
\usepackage{subfigure}

\usepackage{natbib}
\bibpunct[, ]{(}{)}{;}{a}{}{,}

\begin{document}

\articletype{RESEARCH ARTICLE}

\title{Region-Based Classification of PolSAR Data Using Radial Basis Kernel Functions With Stochastic Distances}

\author{
\name{
Rog\'{e}rio~G.~Negri\textsuperscript{a}\thanks{CONTACT -- R.~G. Negri. Email: rogerio.negri@ict.unesp.br}, 
Alejandro~C.~Frery\textsuperscript{b},
Wagner~B.~Silva\textsuperscript{c},
Tatiana~S.~G.~Mendes\textsuperscript{a} and
Luciano~V.~Dutra\textsuperscript{d}
}
\affil{
\textsuperscript{a}UNESP -- Universidade Estadual Paulista, ICT -- Instituto de Ci\^{e}ncia e Tecnologia, S\~{a}o Jos\'{e} dos Campos, Brazil; \\
\textsuperscript{b}UFAL -- Universidade Federal de Alagoas, LaCCAN -- Laborat\'orio de de Computa\c{c}\~{a}o Cient\'ifica e An\'alise Num\'erica, Macei\'{o}, Brazil; \\
\textsuperscript{c}IME -- Instituto Militar de Engenharia, Se\c{c}\~{a}o de Ensino de Engenharia Cartogr\'{a}fica, Rio de Janeiro, Brazil; \\
\textsuperscript{d}INPE -- Instituto Nacional de Pesquisas Espaciais, DPI -- Divis\~{a}o de Processamento de Imagens, S\~{a}o Jos\'{e} dos Campos, Brazil.
}
}

\maketitle

\begin{abstract}

Region-based classification of PolSAR data can be effectively performed by seeking for the assignment that minimizes a distance between prototypes and segments.
\citet{SilvaEA2013} used stochastic distances between complex multivariate Wishart models which, differently from other measures, are computationally tractable.
In this work we assess the robustness of such approach with respect to errors in the training stage, and propose an extension that alleviates such problems.
We introduce robustness in the process by incorporating a combination of radial basis kernel functions and stochastic distances with Support Vector Machines (SVM). 
We consider several stochastic distances between Wishart: Bhatacharyya, Kullback-Leibler, Chi-Square, R\'{e}nyi, and Hellinger.
We perform two case studies with PolSAR images, both simulated and from actual sensors, and different classification scenarios to compare the performance of Minimum Distance and SVM classification frameworks.
With this, we model the situation of imperfect training samples.
We show that SVM with the proposed kernel functions achieves better performance with respect to Minimum Distance, at the expense of more computational resources and the need of parameter tuning.
Code and data are provided for reproducibility.

\end{abstract}

\begin{keywords}
PolSAR; image classification; stochastic distance; Minimum Distance Classifier; SVM
\end{keywords}

\section{Introduction}\label{intro}

The availability of Polarimetric Synthetic Aperture Radar (PolSAR) sensors has increased as a consequence of the technological advances in Remote Sensing.
Compared to conventional SAR sensors, PolSAR is able to acquire the amplitude, phase, and orientation of the electromagnetic waves reflected from targets in different transmission and reception polarizations.
The use of these images is challenging, among other reasons, because of the data structure (complex matrices in each pixel), their properties (the Gaussian additive noise is not valid), and the signal-to-noise ratio (typically very low).
Bearing these characteristics in mind, several specific methods have been developed for PolSAR image processing and classification.

PolSAR image classification has been intensively investigated. 
The notion that more information on the same area leads to better classification is intuitive. 
\citet{LeeGrunesKwok1994} and \citet{CloudePottier1997} proposed two pioneer approaches for PolSAR data classification.
While the former develops a pixel-based method based on the Maximum Likelihood Classifier under the Complex Multivariate Wishart distribution, 
the latter is an unsupervised classification through the eigenvalue analysis of coherency matrices.

Several PolSAR data classification approaches have been proposed since then.
\citet{FreryEA2007} used specific probability density functions for PolSAR intensity data in the classic Maximum Likelihood Classifier method. 
\citet{ErsahinEA2010} based their approach for segmentation and classification of PolSAR data on spectral graph partitioning. 
\citet{DuEA2014} used a Kernel Extreme Learning Machine with multiple polarimetric and spatial features.
\citet{TaoEA2015} proposed a feature extraction method based on Independent Component Analysis and tensor decomposition.
Recently, \citet{HuoEA2017} proposed a semi-supervised method able to learn even when the training data quality and quantity are both poor.

\citet{SilvaEA2013} investigated the use of stochastic distances between Complex Multivariate Wishart distributions on a region-based approach. 
The performance of using the Kullback-Leibler, Bhattacharyya, Hellinger, R\'{e}nyi, and Chi-Square stochastic distances was assessed, providing evidence that such leads to better results when compared to the Maximum Likelihood Classifier using the Complex Multivariate Wishart distribution for pixel-based PolSAR classification as proposed in \citet{LeeGrunesKwok1994}. 
Furthermore, the Kullback-Leibler, Bhattacharyya, and R\'{e}nyi distances are more indicated than Hellinger and Chi-Square.
\citet{ClassificationPolSARGeodesic} proposed projecting PolSAR data onto Kennaugh matrices, and then computing the geodesic distance between elementary targets and the observed information.

\citet{negri2016kmeans} presented a new version of K-Means algorithm for region-based classification of PolSAR data by using a stochastic distance between Complex Multivariate Wishart models and a hypothesis test derived from this kind of measure.
Similarly, \citet{NegriEA2016b} verified the use of Bhattacharyya distance with a Support Vector Machine (SVM) through kernel functions for region-based classification of SAR data.
However, to the best of the authors' knowledge, there is still room for investigating other distances and specific distributions for PolSAR images.
Also, the influence of errors on the training data has received little attention in the literature.

This study aims at analyzing the use of a number of stochastic distances as inputs for an SVM with kernel functions for PolSAR region-based classification. 
Additionally, it presents comparisons with the Minimum Distance Classifier framework investigated by \citet{SilvaEA2013}. 
Such comparisons are conducted on PolSAR images, both simulated and from operational sensors, and different classification scenarios.
Such scenarios include an analysis of the often encountered situation of using imperfect, i.e. contaminated, training samples.

The remainder of this article is organized as follows.
Fundamental concepts regarding statistical PolSAR modelling, and the use of stochastic distances for region-based classification are presented in Section~\ref{Theo}. 
In Section~\ref{experiments}, these concepts are applied in two case studies. 
Conclusions are presented in Section~\ref{concl}.

\section{Statistical region-based PolSAR classification}\label{Theo}

\subsection{Statistical PolSAR modelling}\label{PolSARModelling}

The backscatter signal measured by a PolSAR sensor can be represented through the complex scattering vector $\mathbf{z}^{T} = \left( S_{hh} \ S_{hv} \ S_{vh} \ S_{vv} \right)$. 
Each component of $\mathbf{z}$ is a complex number that carries the amplitude and phase of a polarization combination.
The polarizations are indicated by the subscripts in $\mathbf{z}$, where, for example, $hv$ denotes the signal recorded with vertical polarization from a signal initially emitted with horizontal polarization. 
Considering the reciprocity of an atmospheric medium, which makes $S_{hv}$ similar to $S_{vh}$, the scattering vector is simplified to $\mathbf{z}^{T} = \left(S_{hh} \ S_{hv} \ S_{vv} \right)$ \citep{FreryEA2007}. 

An $N$-looks covariance matrix is the average of $N$ backscatter measurements in a neighborhood:
\begin{equation}\label{MC}
\mathbf{Z} = \frac{1}{N}\sum_{\ell=1}^{N}  \mathbf{z}_{\ell} \mathbf{z}_{\ell}^{\star T} =
\begin{pmatrix}
{Z}_{hh} & {Z}_{hhhv} & {Z}_{hhvv} \\
{Z}_{hhhv}^{\star} & {Z}_{hv} & {Z}_{hvvv} \\
{Z}_{hhvv}^{\star} & {Z}_{hvvv}^{\star} & {Z}_{vv}
\end{pmatrix},
\end{equation}
where $\star$ and $T$ represent the conjugate and transposed operators, respectively. 
The diagonal elements of  $\mathbf{Z}$ are nonnegative numbers that represent the intensity of the signal measured on a specific polarization.

Assuming that $\mathbf{z}$ follows a zero-mean Complex Gaussian distribution~\citep[cf.][]{Goodman1963}, it is possible to obtain the distribution of $\mathbf{Z}$, the scaled Complex Multivariate Wishart law, which is characterized by the following probability density function:

\begin{equation} \label{WCM}
f\left( \mathbf{Z};N, \mathbf{\Sigma} \right) = \frac{N^{3N} |\mathbf{Z}|^{N-3} \exp\{-N Tr\left( \mathbf{\Sigma}^{-1} \mathbf{Z} \right)\}}{ |\mathbf{\Sigma}|^{N} \Gamma_{3}(N)},
\end{equation} 
where $\Gamma_{3}(N) = \pi^{3}\prod_{i=0}^{2}\Gamma(N - i), N \geq 3$.

The parameters $N$ and $\mathbf{\Sigma}$ are the number of looks and the target mean covariance matrix.
The determinant, inversion and trace operators are denoted $|\cdot|$, $(\cdot)^{-1}$ and $Tr\left( \cdot \right)$, respectively.

The scaled Complex Multivariate Wishart model is valid in textureless areas.
Target variability can be included by one or more additional parameters; the reader is referred to the work by \citet{SurveyStatisticalPolSAR} for a comprehensive survey of models for PolSAR data.

\subsection{Stochastic distances between Complex Multivariate Wishart distributions}\label{stochWishart}

A divergence is a measure of the difficulty of discriminating between two models.
\citet{Csiszar1967} proposed the $\phi$ divergence family, providing a formally organized framework to analytically obtain divergence measures between distributions. 
\citet{Salicru1994} proposed a more general class of divergences, the $\hphi$ divergence, through the adoption of an additional function ($h$).

Consider the random variables $X$ and $Y$ defined on the same support $\mathcal{X}$ with distributions characterized by the densities $f_{X}(\mathbf{x};\bm{\mathbf{\theta}}_{1})$ and $f_{Y}(\mathbf{x};{\bm\theta}_{2})$, respectively, where ${\bm\theta}_{1}$ and ${\bm\theta}_{2}$ are parameters.
The $\hphi$ divergence between $X$ and $Y$ is given by:
\begin{equation} 
\label{eqdivhf}
d^{h}_{\phi}(X,Y)= h \left(\int_{\mathbf{x} \in \mathcal{X}}\phi\left(\frac{f_{X}(\mathbf{x};\bm{\theta}_{1})}{f_{Y}(\mathbf{x};\bm{\theta}_{2})}\right)f_{Y}(\mathbf{x};\bm{\theta}_{2})d\mathbf{x} \right),
\end{equation}
where $\phi\colon(0,\infty)\rightarrow[0,\infty)$ is a convex function and $h\colon(0,\infty)\rightarrow[0,\infty)$ is a strictly increasing function with $h(0)=0$ and $h'(v) > 0$ for all $v \in (0,\infty)$.
Several well-known divergence measures can be obtained by choosing $h$ and $\phi$, but they are not necessarily symmetric.

The symmetrization $\displaystyle \text{D}(X,Y) = (d^{h}_{\phi}(X,Y) + d^{h}_{\phi}(Y,X))/2$ allows to obtain distances measures from any divergence $d^{h}_{\phi}$. This leads to the following properties:
\begin{enumerate}
\item {\bfseries Non-negativity: }$\textrm{D}(X,Y) \geq 0$;
\item {\bfseries Identity (of indiscernible): }$\textrm{D}(X,Y) = 0 \Leftrightarrow X = Y$;
\item {\bfseries Symmetry: }$\textrm{D}(X,Y) = \textrm{D}(Y,X)$.
\end{enumerate}
Additionally, if a distance has the following property, then it is a metric:
\begin{enumerate}
\setcounter{enumi}{3}
\item {\bfseries Triangle inequality: }$\textrm{D}(X,Z) \leq \textrm{D}(X,Y) + \textrm{D}(Y,Z)$,
\end{enumerate}
where $X$, $Y$ and $Z$ have the same support $\mathcal{X}$. 
It is noteworthy that Bathacharrya, Kullback-Leibler, R\'{e}nyi, Hellinger and Chi-Square distances are not metrics since the triangle inequality is not fulfilled.

\citet{NascimentoEA2010} computed the Bathacharrya, Kullback-Leibler, R\'{e}nyi (of order $\beta$), Hellinger, Jensen-Shannon, Arithmetic-Geometric, Triangular and Harmonic Mean stochastic distances between $\mathcal G^0$ distributions. 
These distances were successfully used to evaluate contrast differences among regions in intensity Synthetic Aperture Radar (SAR) images.

\citet{FreryEA2014} developed analytic expressions for the first four aforementioned stochastic distances between scaled Complex Multivariate Wishart distributions.
Previously, the Chi-Square stochastic distance between Complex Multivariate Wishart distributions was presented in \citet{FreryEA2011}.
These distances were used by~\citet{SilvaEA2013} in PolSAR imagery classification by a minimum distance criterion.

Let $X$ and $Y$ be two random variables modelled by scaled Complex Multivariate Wishart distributions $X\sim \mathcal{W}(\mathbf{\Sigma}_1,N)$ and $Y\sim \mathcal{W}(\mathbf{\Sigma}_2,N)$. 
The Bathacharrya, Kullback-Leibler, R\'{e}nyi (order $0<\beta<1$) and Hellinger stochastic distances expressions between scaled Complex Multivariate Wishart distributions, according to~\citet{FreryEA2014}, are given by:
\begin{align}
{\textrm{D}}_\textrm{B}(X,Y) &= N \left[\frac{\log|\mathbf{\Sigma}_1|+\log|\mathbf{\Sigma}_2|}{2}-\log\left|\left(\frac{\mathbf{\Sigma}_1^{-1}+\mathbf{\Sigma}_2^{-1}}{2}\right)^{-1}\right|\right],
\label{distB}\\
{\textrm{D}}_\textrm{K}(X,Y) & = N \left[\frac{Tr(\mathbf{\Sigma}_1^{-1}\mathbf{\Sigma}_2+\mathbf{\Sigma}_2^{-1} \mathbf{\Sigma}_1)}{2} - 3 \right],
\label{distK}\\
{\textrm{D}}_\textrm{R}^{\beta}(X,Y) & = \frac{\log 2}{1-\beta}+\frac{1}{\beta-1}\log \left\lbrace \left[|\mathbf{\Sigma}_1|^{-\beta}|\mathbf{\Sigma}_2|^{\beta-1}|(\beta \mathbf{\Sigma}_1^{-1}+(1-\beta) \mathbf{\Sigma}_2^{-1})^{-1}|\right]^N + \right.\nonumber \\
 & + \left. \left[| \mathbf{\Sigma}_1|^{\beta-1}| \mathbf{\Sigma}_2|^{-\beta}|(\beta \mathbf{\Sigma}_2^{-1}+(1-\beta) \mathbf{\Sigma}_1^{-1})^{-1}|\right]^N \right\rbrace, \label{distR}\\
{\textrm{D}}_\textrm{H}(X,Y) & = 1-\left[\frac{\left| 2 ({\mathbf{\Sigma}_1^{-1}+\mathbf{\Sigma}_2^{-1}})^{-1} \right| }{\sqrt{|\mathbf{\Sigma}_1||\mathbf{\Sigma}_2|}}\right]^N,
\label{distH}\\
{\textrm{D}}_\textrm{C}(X,Y) & =   \left( \frac{|\mathbf{\Sigma}_1|}{|\mathbf{\Sigma}_2|^2} \textrm{abs}\left( \left| \left( 2\mathbf{\Sigma}_{2}^{-1} - \mathbf{\Sigma}_{1}^{-1} \right)^{-1} \right| \right) \right)^{N} + \nonumber \\
 & + \left( \frac{|\mathbf{\Sigma}_{2}|}{|\mathbf{\Sigma}_{1}|^{2}} \textrm{abs} \left( \left| \left( 2\mathbf{\Sigma}_{1}^{-1} - \mathbf{\Sigma}_{2}^{-1} \right)^{-1} \right| \right) \right)^{N} - 2 ,
\label{distC}
\end{align}
where $\textrm{abs}(\cdot)$ returns the modulus of a real number.


\subsection{Region-based PolSAR image classification}\label{rbcPolSAR}

Let $\mathcal{I}$ be an image defined on the grid $\mathcal{S} \subset \mathbbm{N}^{2}$ whose pixels are elements of the attribute space $\mathcal{X}$. 
We use the notation $\mathcal{I}(s)=\mathbf{x}$ to represent that a pixel $s \in \mathcal{S}$ of $\mathcal{I}$ has attribute $\mathbf{x} \in \mathcal{X}$.
The image support can be partitioned in $r\geq 1$ disjoint subsets $\mathcal{R}_{i} \subset \mathcal{S}, \ i = 1, \ldots, r$, such that $\cup_{i=1}^{r} \mathcal{R}_{i} = \mathcal{S}$.

A region-based classification process consists of associating a class $\omega_{j}$ from a set of $c$ possible classes $\Omega$ to all the pixels that comprise the region $\mathcal{R}_{i}$. 
A supervised region-based decision rule is built trough information available from a set of labeled regions $\mathcal{D} = \left\lbrace \left( \mathcal{R}_{k},\omega_{j} \right) \in \mathcal{S} \times \Omega : k = 1, \ldots, s; j = 1, \ldots, c \right\rbrace$. 
The notation $\left( \mathcal{R}_{k},\omega_{j} \right)$ indicates that  $\mathcal{R}_{k}$ is assigned to the class $\omega_{j}$.

\citet{SilvaEA2013} adopt the Minimum Distance Classifier framework using stochastic distances as a measure to compare the similarity between classes and unlabeled regions. 
We refer to this method as Minimum Stochastic Distance Classifier ({MSDC}). 
The pixel values in an unlabeled region are used to estimate a probability distribution. 
This region is then assigned to the closest class in distribution, according to an stochastic distance. 
The class distributions are modeled based on information from $\mathcal{D}$. 

Formally, let $\mathcal{R}_{i}$ be an unlabeled region and let $\textrm{D}(\widehat{f}_{\mathcal{R}_{i}},\widehat{f}_{\omega_{j}})$ be a stochastic distance between distributions estimated from the attributes of the pixels in $\mathcal{R}_{i}$ and the class $\omega_{j}$.
An assignment $(\mathcal{R}_{i},\omega_{j})$ is made when the following rule is satisfied:
\begin{equation} \label{MDC}
(\mathcal{R}_{i},\omega_{j}) \Leftrightarrow j = \underset{j=1,\ldots,c}{\arg \textrm{min}} \ \textrm{D}(\widehat{f}_{\mathcal{R}_{i}}, \widehat{f}_{\omega_{j}}).
\end{equation}
In Equation \eqref{MDC}, $\widehat{f}_{\omega_{j}}$ is estimated with all the pixels assigned to $\omega_{j}$ in $\mathcal{D}$.

Other methods can be adopted to perform region-based classification beyond the {MSDC} as, for instance, Support Vector Machines (SVMs). 
SVMs have received great attention because of their excellent generalization ability, their independence of data distribution and their robustness with respect to the Hughe's phenomenon \citep{BruzzonePersello2009}.

The use of {\itshape kernel functions}, $K\colon\mathcal{X}^{2} \rightarrow \mathbbm{R}$, is a common strategy to improve SVM classification performance on nonlinearly separable patterns. 
Kernel functions also allow the application of SVMs in problems where patterns cannot be represented as vectors.


A kernel function $K\colon\mathcal{X}^{2} \rightarrow \mathbbm{R}$ is symmetric and conforms to the Mercer theorem conditions \citep{TK2008}. 
However, as such verification may be not trivial there are alternative ways to develop such functions.
For example, adopting the radial basis function model \citep{Scholkopf2002}:
\begin{equation}\label{modker}
K(\mathbf{x}_{u},\mathbf{x}_{v}) = g\left( \widetilde m \left( \mathbf{x}_{u},\mathbf{x}_{v} \right)\right),
\end{equation}
where $g\colon\mathbbm{R} \rightarrow \mathbbm{R}$ is a strictly positive real function and $\widetilde m\colon\mathcal{X}^{2} \rightarrow \mathbbm{R}$ is a metric.

Equation~\eqref{modker} gives a hint to develop suitable kernel functions for PolSAR region-based image classification. 
A sensible choice for $g$ is the negative exponential function, and
$m$, as a measure of similarity between the input data, a metric based on stochastic distances between distributions. 
It is noteworthy that equations~\eqref{distB} to~\eqref{distC} will not produce valid kernel functions since they do not attain the triangle inequality.

However, if $\textrm{D}$ is a stochastic distance with $\tau \in \mathbbm{R}_{+}$ such that $\textrm{D}(\mathcal{R}_{u} , \mathcal{R}_{v}) \leq \tau$ for $u,v = 1,\ldots, r$, the following expression provides a metric:
\begin{equation}\label{transDistMet}
m(\mathcal{R}_{u},\mathcal{R}_{v}) = \left\lbrace
\begin{array}{ll}
0 & \textrm{if} \ \mathcal{R}_{u} = \mathcal{R}_{v}, \\
\textrm{D}(\mathcal{R}_{u},\mathcal{R}_{v}) + \tau & \textrm{if} \ \mathcal{R}_{u} \neq \mathcal{R}_{v}.
\end{array}
\right.
\end{equation}

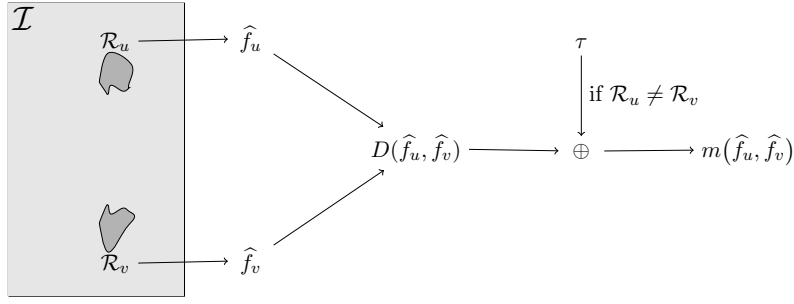
\begin{figure}[hbt]
\begin{center}
\begin{adjustbox}{max width=0.75\textwidth}
\begin{tikzpicture}
   \draw [line width=0.25mm] (-7.5,-2.5) -- (-4.5,-2.5) -- (-4.5,2.5) -- (-7.5,2.5) -- (-7.5,-2.5);
   \fill [black!10] (-7.5,-2.5) rectangle (-4.5,2.5);  
   
   \draw  [fill=black!30] plot[smooth, tension=.7, line width=0.5mm] coordinates {(-5.95,-1.25) (-5.8,-1.75)  (-5.5,-1.35) (-5.35,-1.15) (-5.45,-1.05) (-5.55,-1.0) (-5.75,-1.1) (-5.8,-0.95) (-5.85,-1.0) (-5.95,-1.25)};  

   \draw  [fill=black!30] plot[smooth, tension=.7, black!50, line width=0.5mm] coordinates {(-5.95,1.15) (-5.8,1.65)  (-5.4,1.45) (-5.45,1.1) (-5.45,1.05) (-5.55,1.0) (-5.75,1.1) (-5.8,0.95) (-5.85,1.0) (-5.95,1.15)};   
  
   \node [scale=1.5] at (-7.25,2.25) {$\mathcal{I}$};  
 
   \matrix (m) [matrix of math nodes,row sep=3em , column sep=4em , minimum width=2em]
   {
     \mathcal{R}_{u} &    \widehat{f}_{u}    &               						   &  \tau     &     										  	     \\
           		     &           			 &   D(\widehat{f}_{u},\widehat{f}_{v})    &  \oplus   &  m\big( \widehat{f}_{u} , \widehat{f}_{v} \big)  \\
     \mathcal{R}_{v} &     \widehat{f}_{v}   &               						   &           &     									 			 \\
   };

   \path[=stealth, ->]
      (m-1-1) edge node [left] {} (m-1-2)
      (m-3-1) edge node [left] {} (m-3-2)

      (m-1-2) edge node [left] {} (m-2-3)
      (m-3-2) edge node [left] {} (m-2-3)
    
      (m-2-3) edge node [left] {} (m-2-4)
    
      (m-1-4) edge node [left] [right] {if $\mathcal{R}_{u} \neq \mathcal{R}_{v}$} (m-2-4)
    
      (m-2-4) edge node [left] {} (m-2-5);
\end{tikzpicture}
\end{adjustbox}
\end{center}
\caption{Schema of the construction of a kernel function $K$ between regions $\mathcal R_u$ and $\mathcal R_v$.}\label{fig:schema}
\end{figure}

The identity property of $m$ stems from~\eqref{transDistMet}.
Non-negativity and symmetry are inherited from $\textrm{D}$, which is a distance, since adding a positive constant will not invalidate such proprieties. 
Finally, the triangle inequality is fulfilled since the following relations are satisfied:

\begin{gather*}
m(\mathcal{R}_{u},\mathcal{R}_{v}) + m(\mathcal{R}_{v},\mathcal{R}_{w}) \geq m(\mathcal{R}_{u},\mathcal{R}_{w}) \Leftrightarrow \\
\Leftrightarrow \textrm{D}(\mathcal{R}_{u},\mathcal{R}_{v}) + \tau + \textrm{D}(\mathcal{R}_{v},\mathcal{R}_{w}) + \tau \geq \textrm{D}(\mathcal{R}_{u},\mathcal{R}_{w}) + \tau \Leftrightarrow \\
\Leftrightarrow \textrm{D}(\mathcal{R}_{u},\mathcal{R}_{v}) + \textrm{D}(\mathcal{R}_{v},\mathcal{R}_{w}) + \tau \geq \textrm{D}(\mathcal{R}_{u},\mathcal{R}_{w}),
\end{gather*}
once we have that $\textrm{D}(\mathcal{R}_{u},\mathcal{R}_{w}) \leq \tau$.
The constant $\tau$ can be chosen as the largest distance measured by $\textrm{D}$ in the classification problem.
Figure~\ref{fig:schema} illustrates this procedure.

We can thus define kernel functions for PolSAR region-based image classification using the model from Equation~\eqref{modker}, the expression from Equation~\eqref{transDistMet} and considering the substitution of $\textrm{D}$ by a stochastic distance given by Equations~\eqref{distB} to~\eqref{distC}:
\begin{equation}
K(\mathcal R_{u},\mathcal R_{v}) = e^{-\gamma  m\left( \mathcal R_{u},\mathcal R_{v} \right) },
\label{geralModKernel}
\end{equation}
where $\gamma \in \mathbbm{R}_{+}$ is a user-adjusted parameter. 
We denote the kernel functions obtained from the Bathacharrya, Kullback-Leiber, Hellinger, R\'{e}nyi e Chi-Square distances as ${\textrm{K}}_\textrm{B}$, ${\textrm{K}}_\textrm{K}$, ${\textrm{K}}_\textrm{H}$, ${\textrm{K}}_\textrm{R}$ and ${\textrm{K}}_\textrm{C}$, respectively.

We make clear that $K(\mathcal R_{u},\mathcal R_{v})$ denotes a kernel function between estimates of distributions $\widehat f_u$ and $\widehat f_v$ computed with the observations in regions $\mathcal R_u$ and $\mathcal R_v$.

\section{Experiments and Results}\label{experiments}

In this section, we present studies of region-based image classification of actual and simulated PolSAR data using the stochastic distances presented in Section~\ref{stochWishart} by both MSDC (i.e., ${\textrm{D}}_\textrm{B}$, ${\textrm{D}}_\textrm{K}$, ${\textrm{D}}_\textrm{H}$, ${\textrm{D}}_\textrm{R}$ and ${\textrm{D}}_\textrm{C}$) and through the kernel functions, defined in Section~\ref{rbcPolSAR} (i.e., ${\textrm{K}}_\textrm{B}$, ${\textrm{K}}_\textrm{K}$, ${\textrm{K}}_\textrm{H}$, ${\textrm{K}}_\textrm{R}$ and ${\textrm{K}}_\textrm{C}$).

The first study (Section~\ref{simCase}) consists of the classification of simulated images generated with basis on parameters estimated from targets in an actual image. 
The second case study (Section~\ref{realcase}) focuses on the classification of the PolSAR image used to extract the parameters. 
The results are compared by accuracy measures and hypothesis tests. 

The scenarios considered in both studies allow a robustness analysis.
Specifically, the results from simulated data are assessed through the overall accuracy, while the actual data are verified with the kappa coefficient of agreement~\citep{Congalton2009}. 

\begin{figure}[hbt]
\centering
{\includegraphics[angle=0, width=14cm]{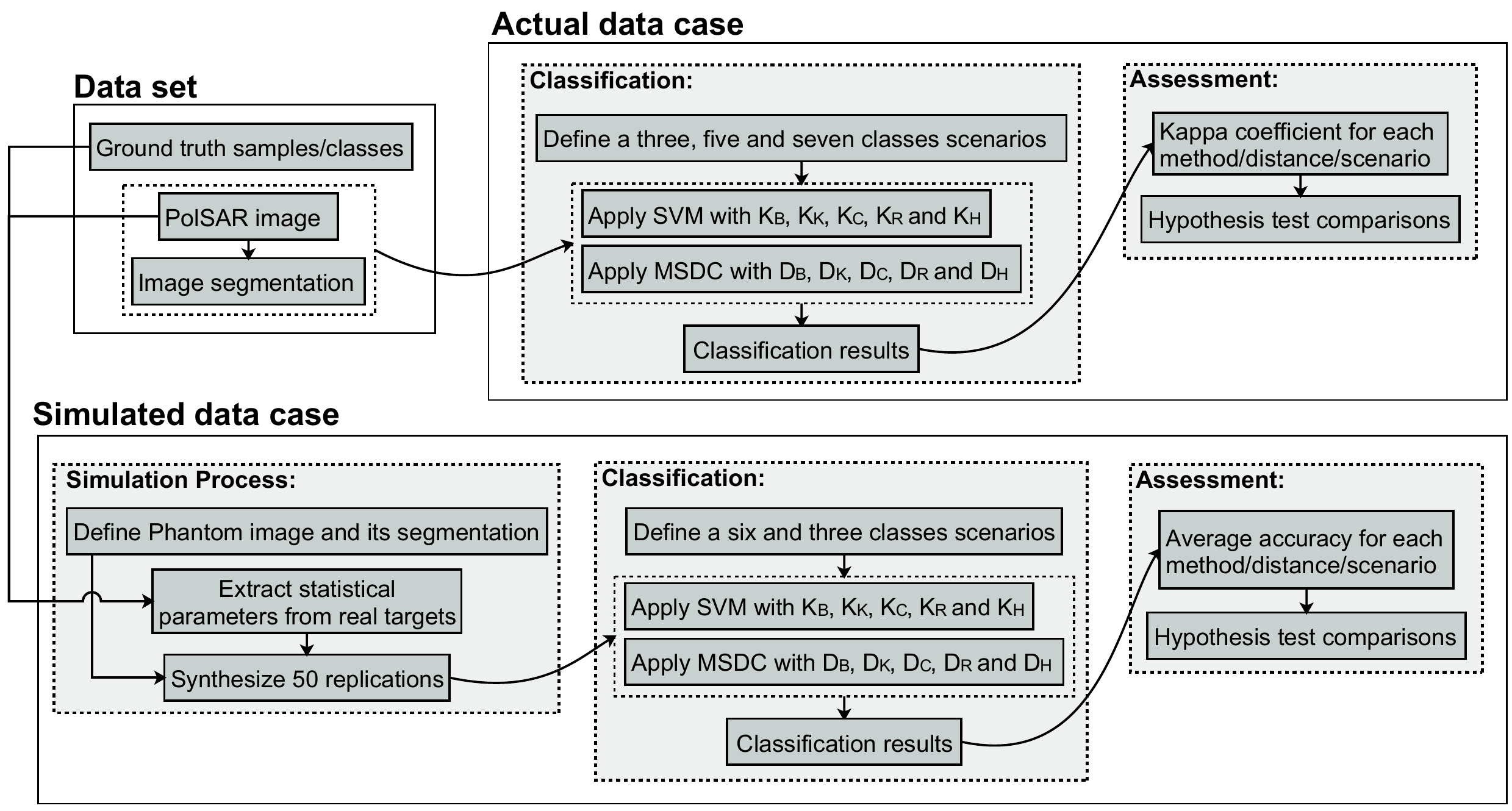} }
\caption{The experiment design overview.}\label{methFig}
\end{figure}

Figure~\ref{methFig} presents an overview of the experiment design, whereas the specifics are discussed in the following sections.

The actual PolSAR data belongs to an image acquired on 
2009 March 13 by the ALOS-PALSAR sensor with approximately
\SI{20x20}{\metre}
resolution after a $3 \times 3$ multi-look process. 
This image, with center near to 
\ang{3;8;19} South and \ang{54;55;26} West, corresponds to a region near the Tapaj\'os National Forest, State of Par\'a, Brazil.

\begin{figure}[hbt]
\centering
\subfigure[Study area location]{\label{AE}\includegraphics[angle=0, width=10cm]{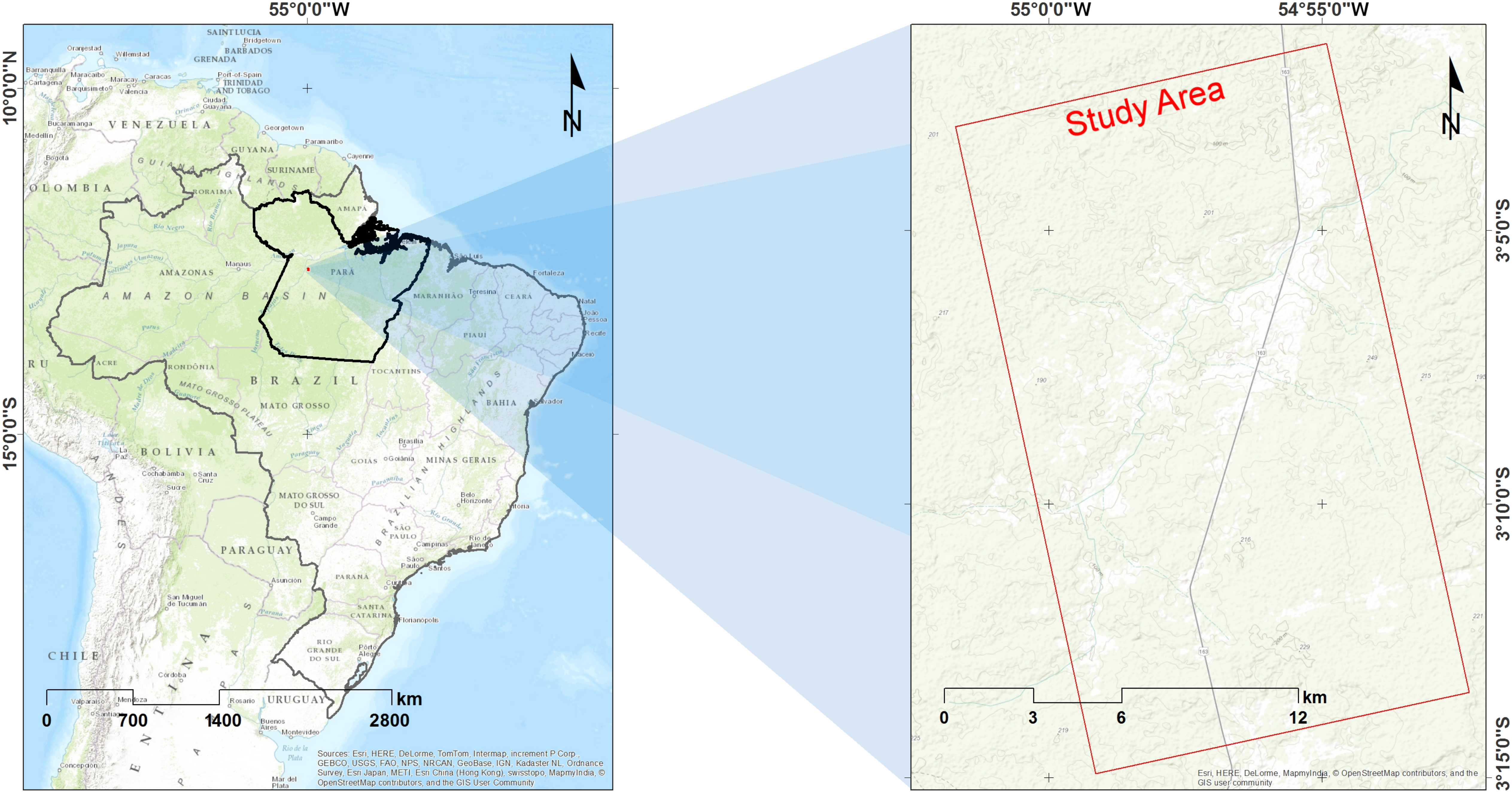} } \\
\subfigure[ALOS-PALSAR image in RGB color composition (HH, HV, VV)]{\label{ImgAE}\includegraphics[width=4.5cm]{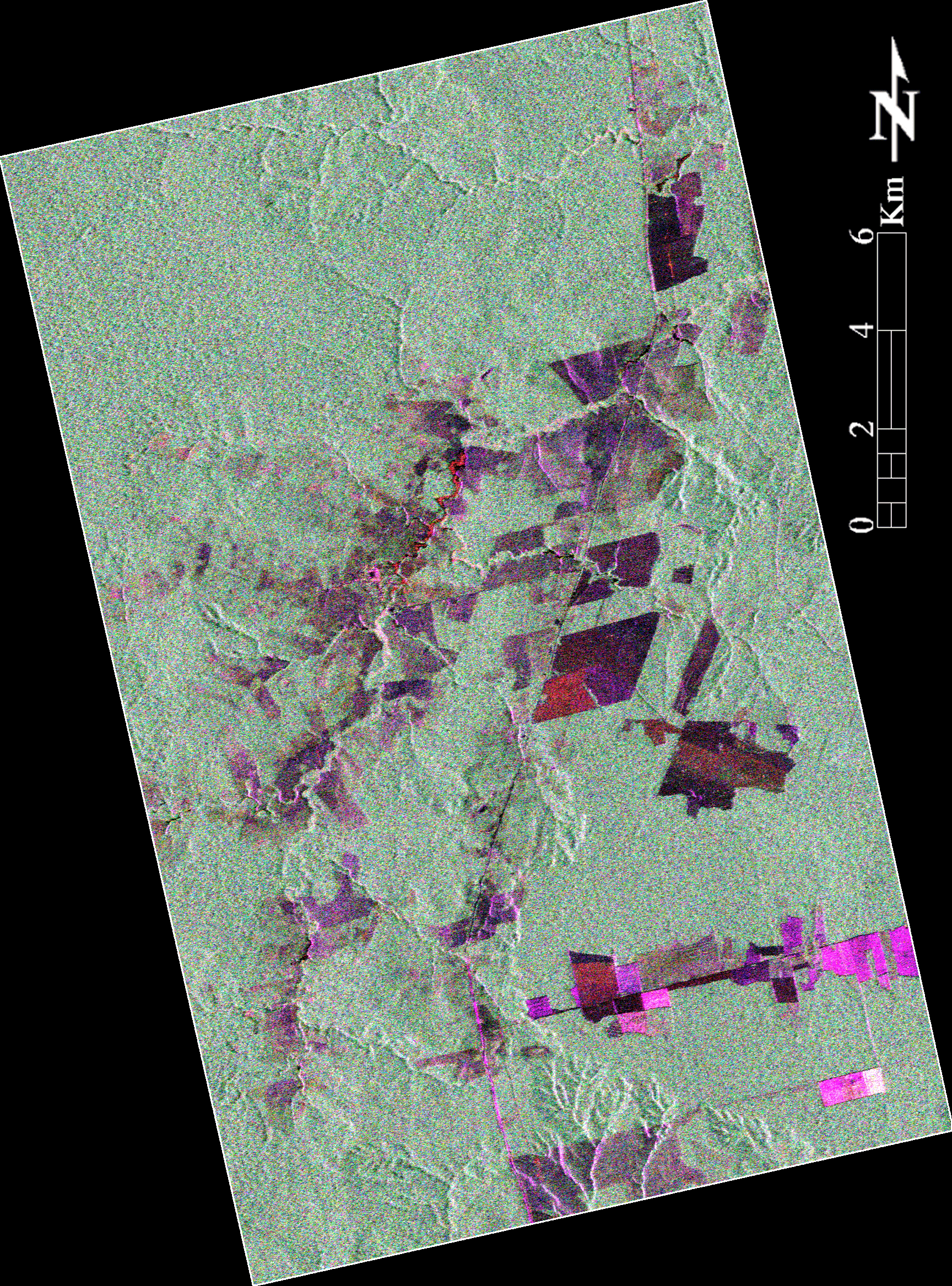} }
\subfigure[LULC samples]{\label{samplesAE}\includegraphics[width=4.5cm]{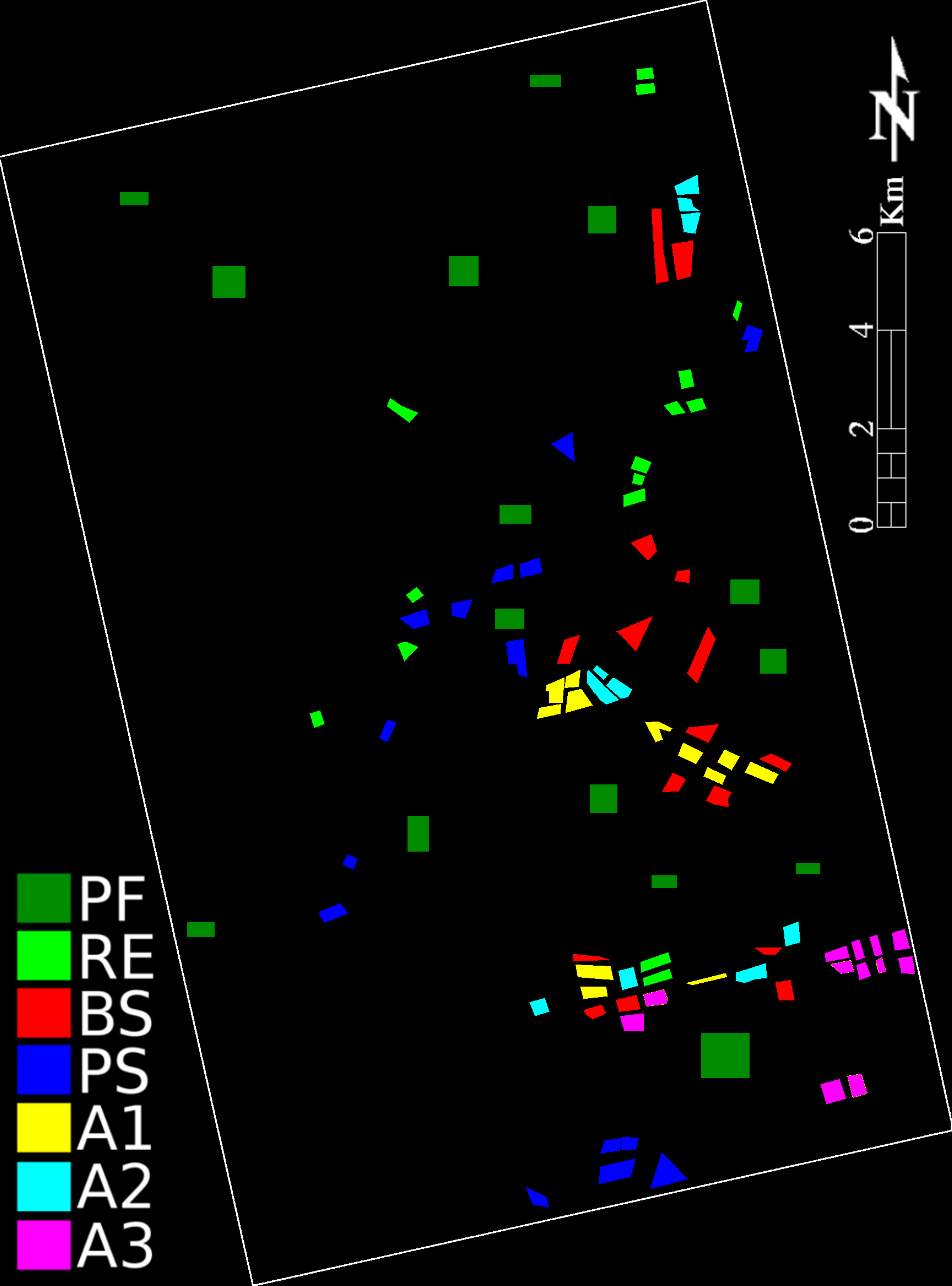}}
\caption{The study area, actual PolSAR image and the spatial distribution of the LULC samples used in the study.}\label{DadosEstudoCaso}
\end{figure}

A field work campaign conducted in September~2009 identified the following land use and land cover (LULC) types: 
Primary Forest~(PF), 
Regeneration~(RE), 
Pasture~(PS), 
Bare Soil~(BS),
and three types of Agriculture (A1, A2 and A3). 
These agricultural classes differ on the crop type or growing stage. 
Figs.~\ref{AE}, \ref{ImgAE} and~\ref{samplesAE} present the study area location, a color composition of the ALOS-PALSAR image and the LULCs, respectively.

The experiments run on a computer with an Intel Core i7 processor, and \SI{16}{\giga\byte} of RAM running the Debian Linux version~8.1 operating system. 
The platform was IDL (Interactive Data Language) version~7.1.
The code is freely available at \url{https://github.com/rogerionegri/SVM-PolSAR}.

\subsection{Classification of Simulated Data}\label{simCase}

\subsubsection{Image Simulation}\label{simProc}

\begin{figure}[hbt!]
\centering
\subfigure[Phantom - Blocks]{\label{phantom}\includegraphics[angle=0, width=12cm]{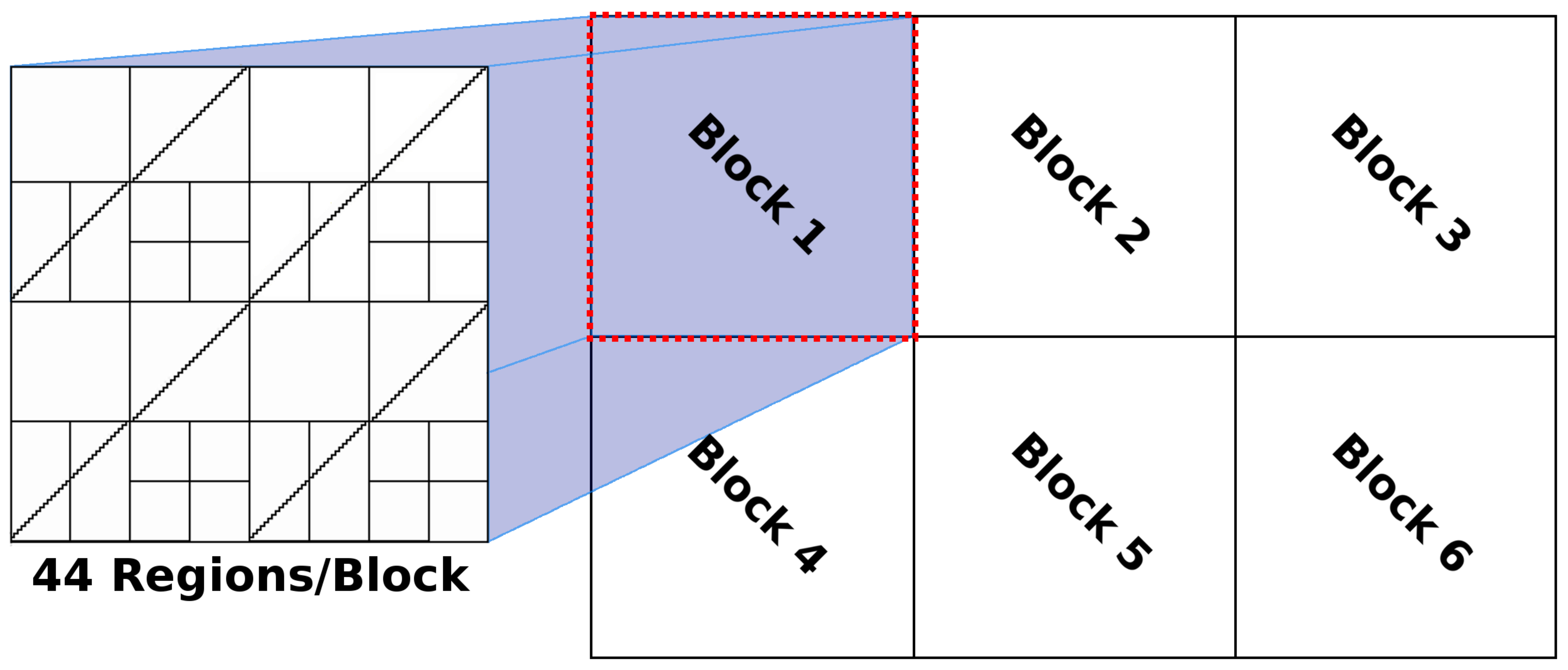} } \\
\subfigure[Simulation example]{\label{exampleSim}\includegraphics[angle=0, width=12cm]{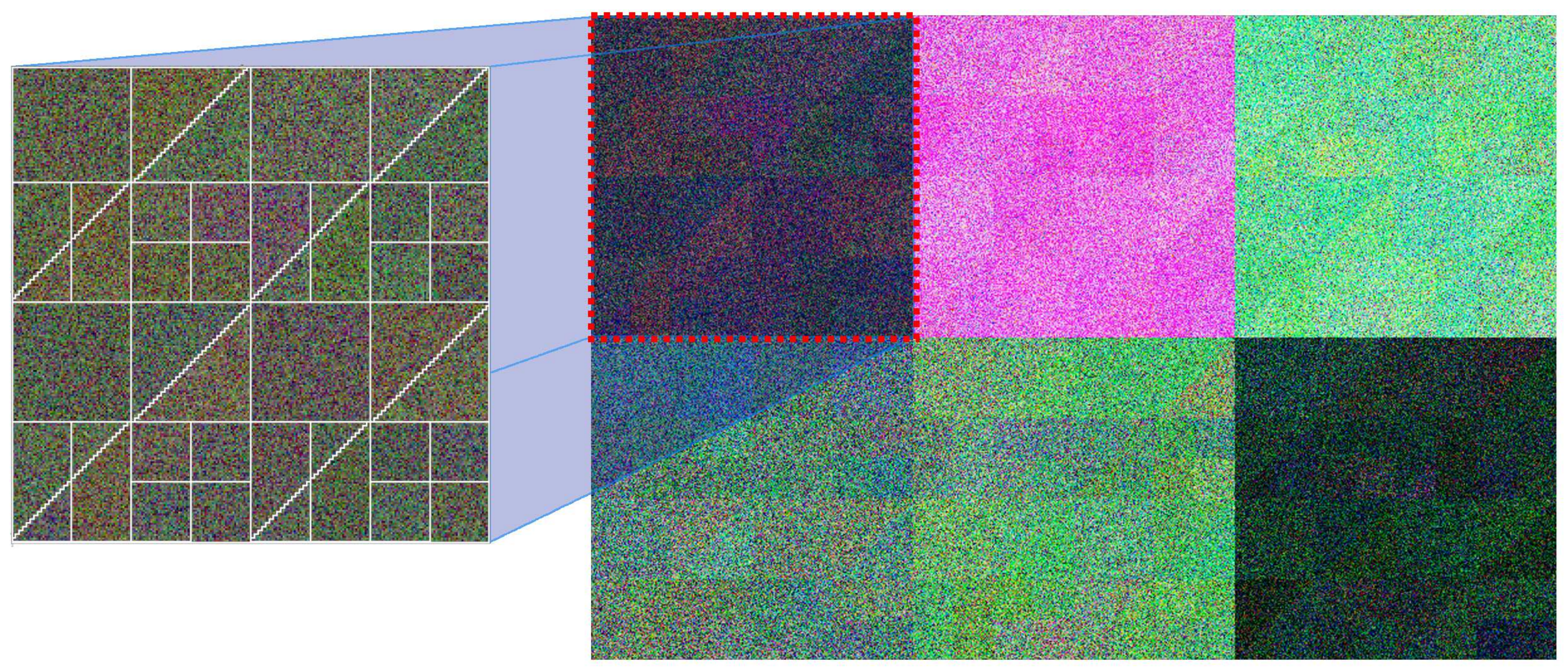} } \\
\subfigure[Mean covariance matrices (squares), and \num{44} perturbed covariance matrices (circles) in semilogarithmic scale.]{\label{scatterSim}\includegraphics[width=12cm]{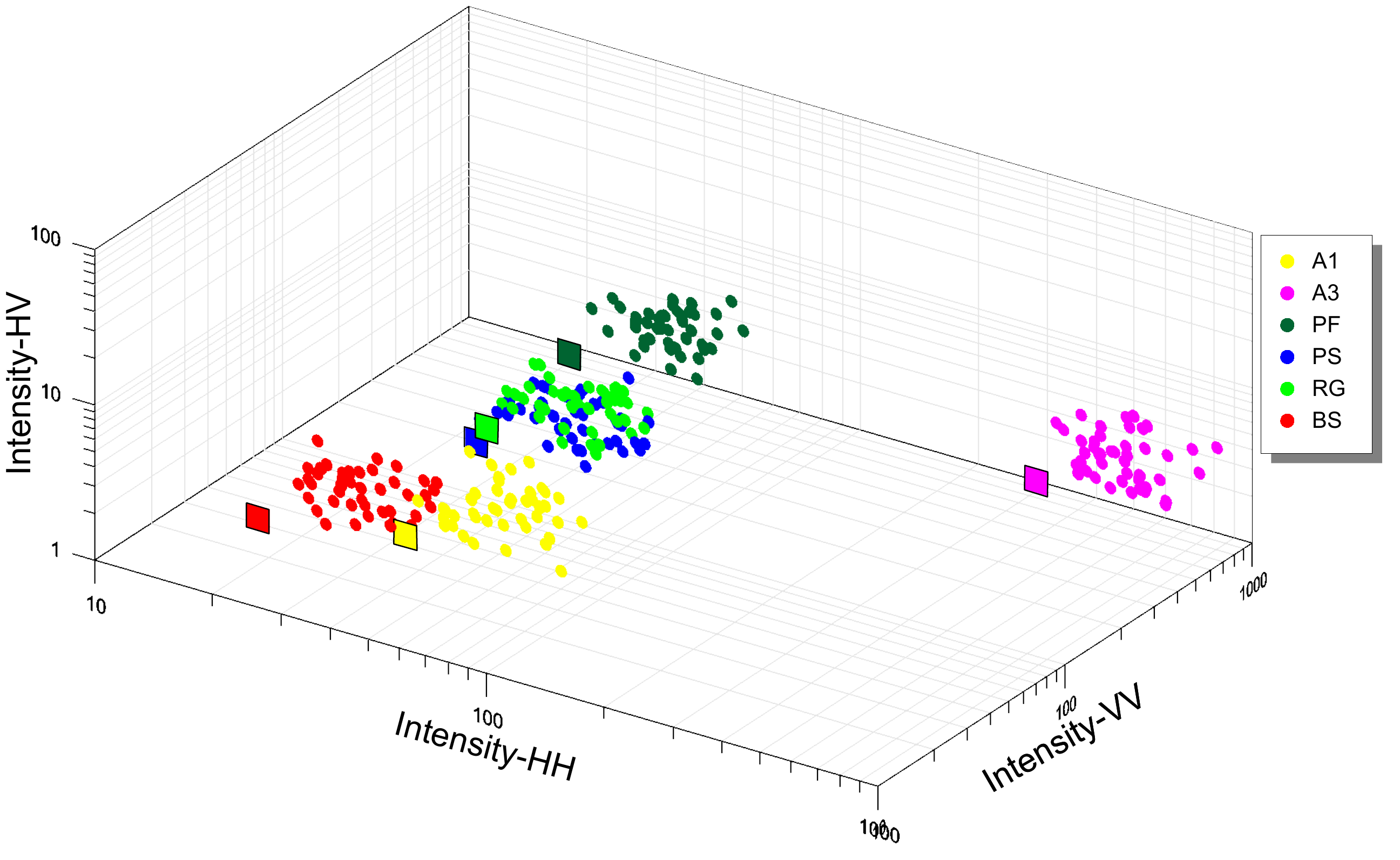} }
\caption{Simulated data.}\label{dadosSimulacao}
\end{figure}

The process adopted for simulated images involves two primary steps. 
The first step consists of defining a ``phantom image'' which is an idealized model for the spatial distribution of classes.
The phantom is formed by six identical blocks of $512 \times 512$ pixels. 
Each block represents a distinct class, which is partitioned into \num{44} segments of different dimension. 
Figure~\ref{phantom} illustrates the phantom image and the structure inside the blocks.

The second step consists of simulating pixels values. 
For this purpose, we adopted a procedure based on \citet{Goodman1963}.
In order to obtain samples from the Wishart distribution with covariance matrix ${\Sigma}$ and $L$ looks, we obtain $L$ independent deviates from scattering vectors following a Complex Multivariate Gaussian distribution with zero mean and covariance matrix $\Theta$:
\begin{equation}\label{scatVecSynth}
\tilde{\mathbf{z}} = \left(z_{1} \ z_{2} \ z_{3}\right) + \left(z_{4} \ z_{5} \ z_{6}\right)\textsf{i}; \ \left(z_{1} \ z_{2} \ z_{3} \ z_{4} \ z_{5} \ z_{6}\right) \sim \mathcal{N}\left( \mathbf{0}, \Theta \right),
\end{equation}
where $\textsf{i}$ represents the imaginary unit and the covariance matrix of the Gaussian law is 
\begin{equation*}
\Theta = \left(
\begin{array}{ccc}
\left[\Re\left( {\Sigma} \right)\right] & \left[ -\Im\left( {\Sigma} \right) \right] \\
\left[ \Im\left( {\Sigma} \right) \right] & \left[ \Re\left( {\Sigma} \right) \right]
\end{array}
\right).
\end{equation*}
We then apply Equation~\eqref{MC} to obtain an observation.
Each class is modeled by a covariance matrix ${\Sigma}$ obtained by averaging observations from the corresponding LULC.

We introduce intra-class variability to describe ``imperfect'' samples and, with this, model plausible errors in the training stage.
\citet{FreryFerreroBustosIJRS_FinalFinal2008} analyzed the effect of such errors in classification, and showed that incorporating context is a way of alleviating such problem.

Consider the (true, unobserved) covariance matrix $\Sigma$ that describes a class.
If it is estimated with perfect samples, i.e., with independent identically distributed observations from the hypothesized distribution, then its maximum likelihood estimate $\widehat\Sigma$ will be as close to $\Sigma$ as the information in the sample permits.
Errors in the training stage may lead to suboptimal estimates in terms of bias and variance.
We model this situation by introducing random perturbations in $\widehat{\Sigma}$, as follows.

The user \textit{believes} each class is characterized by $\Sigma$ but, in fact, the data come from \num{44} slightly different models:
$\Sigma_1,\dots,\Sigma_{44}$.
These models are built by adding random covariance matrices $\Upsilon_1,\dots,\Upsilon_{44}$ to $\widehat\Sigma$, which is obtained estimating from actual data from a single class.
Each perturbation covariance matrix is formed as
$$
\Upsilon_\ell = \left(S_{\ell,hh} \ S_{\ell,hv} \ S_{\ell,vv} \right)^T
\left(S_{\ell,hh} \ S_{\ell,hv} \ S_{\ell,vv} \right)^{\star},1\leq \ell\leq 44,
$$
where $S_{\ell,ij}$ are independent uniform random variables on
$\sqrt{\theta \bar{I}_{ij} (2\sqrt{L})} \cdot (-1,1)$, where $\bar{I}_{ij}$ is the $ij$ diagonal element of $\widehat{\Sigma}$, and $\theta>0$ controls the perturbation.
With this, each mean intensity $\bar{I}_{ij}$ in $\Sigma_\ell$ is a value in $\overline{\widehat{I}}_{ij} \cdot [1,\theta \sigma_{ij}]$, where $\overline{\widehat{I}}_{ij}$ is the (wrongly assumed for the whole block) mean intensity, and $\sigma_{ij}$ is the standard deviation of $\widehat I_{ij}$.

Notice that the correct model for each class would be a mixture of \num{44} distributions, 
but the user will train each class with samples from \num{11} (typically slightly) different laws; cf.\ the colored squares in Fig.~\ref{samplesSim1}.
With this, we assess how the classification techniques perform in the practical situation of using less classes for describing a complex truth, and not collecting samples from all the underlying classes.
Fig.~\ref{scatterSim} shows, in logarithmic scale, the intensity of averaged covariance matrices as squares, and their respective \num{44} perturbed versions.

As can be seen from Fig.~\ref{scatterSim} we consider the following situations:
\begin{itemize}
\item Classes that do not overlap, e.g.\ 
	A1 and A3, A1 and PF, A1 and RG, A3 and all others, PF and BS, PS and BS, RG and BS
\item Classes with some overlap, e.g.\ A1 and PS, A1 and BS,
\item Classes that overlap, e.g.\ PS and RG.
\end{itemize}
This is corroborated by~\eqref{HellingerDistances} below, which shows the Hellinger distances between the classes.

\begin{equation}
\begin{tabular}{c|*5{r}}
$\textrm{D}_{\textrm{H}}$ & A3 & PF & PS & RG & BS\tabularnewline
\thickhline 
A1  & 0.961 & 0.772 & 0.344 & 0.410 & 0.315\tabularnewline
A3 &  & 0.906 & 0.933 & 0.928 & 0.989\tabularnewline
PF & &  & 0.443 & 0.283 & 0.899\tabularnewline
PS & & &  & 0.062 & 0.523\tabularnewline
RG & & & &  & 0.652\tabularnewline
\end{tabular}
\label{HellingerDistances}
\end{equation}

Figure~\ref{exampleSim} presents a simulated image, where it is possible to identify the intra-class variability.
The covariance matrices that plays the rule of $\widehat{\Sigma}$ in the simulation process, estimated from observed LULC samples shown in Figure~\ref{samplesAE}, are presented as Appendix.

\subsubsection{Classification and Results}\label{simClaRes}

A set of fifty images were simulated independently following the procedure described in Section~\ref{simProc}
Each image was then classified using the {MSDC} and {SVM} methods. 
MSDC used the stochastic distances between Complex Multivariate Wishart distributions by plugging Equations~\eqref{distB} to~\eqref{distC} in~\eqref{MDC}.
{SVM} also employed these distances through the kernel functions in~\eqref{transDistMet} and~\eqref{geralModKernel}.

\begin{figure}[hbt!]
\centering
\mbox{
\subfigure[Training samples -- six classes case]{\label{samplesSim1}\includegraphics[angle=0, width=6cm]{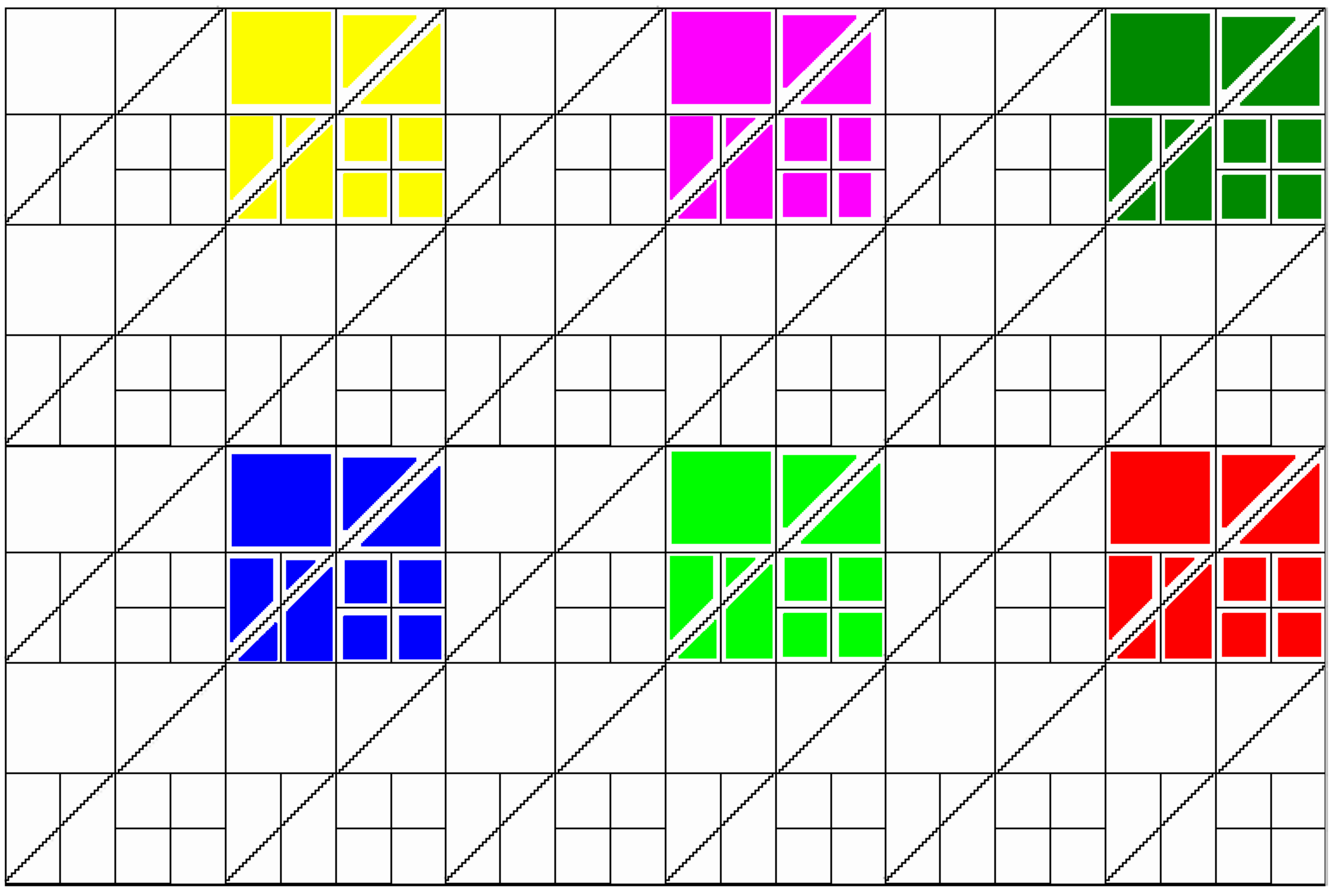} }
\subfigure[Ideal result for six classes]{\label{idealSim1}\includegraphics[angle=0, width=6cm]{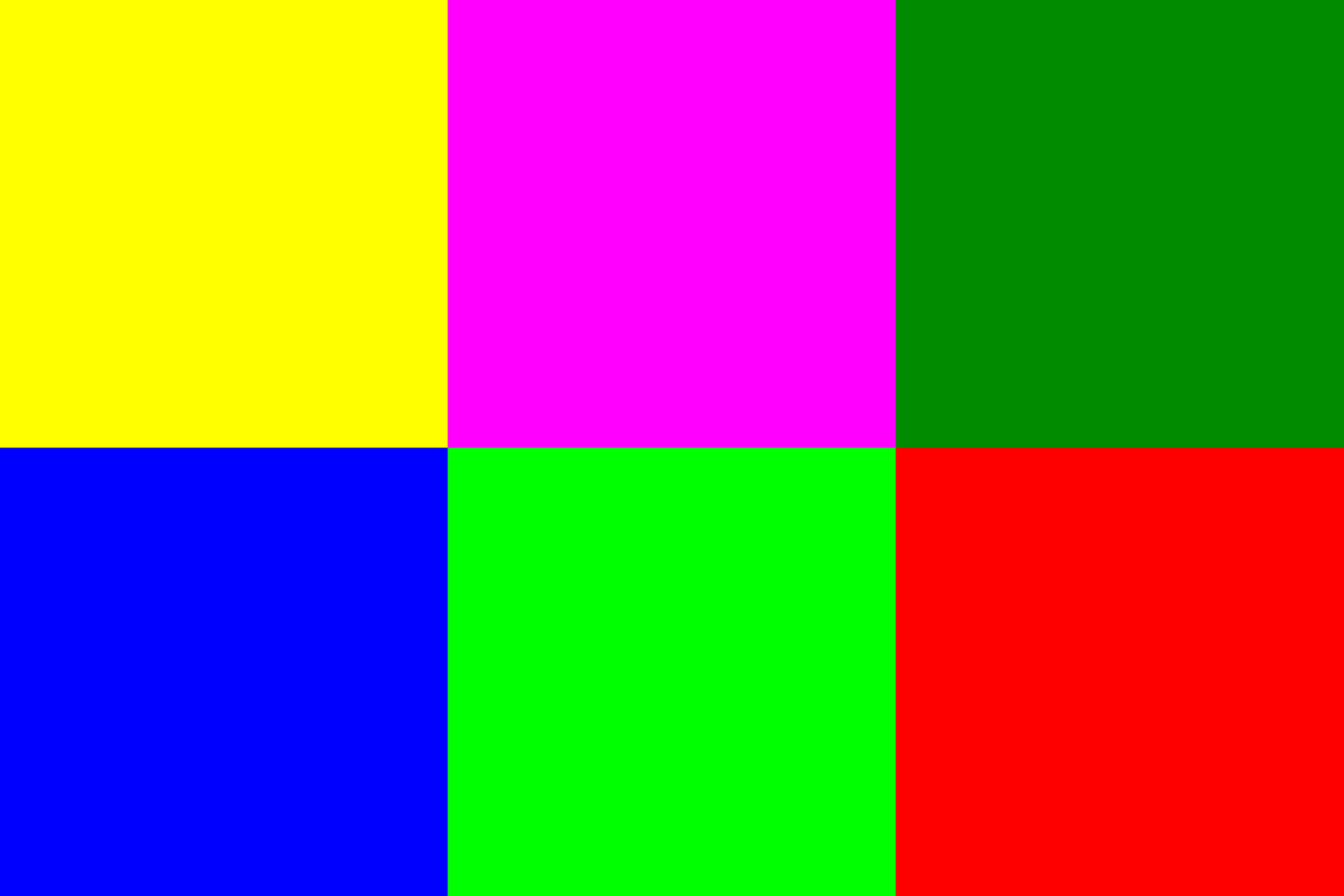} }
} \\
\mbox{
\subfigure[Training samples -- three classes case]{\label{samplesSim2}\includegraphics[angle=0, width=6cm]{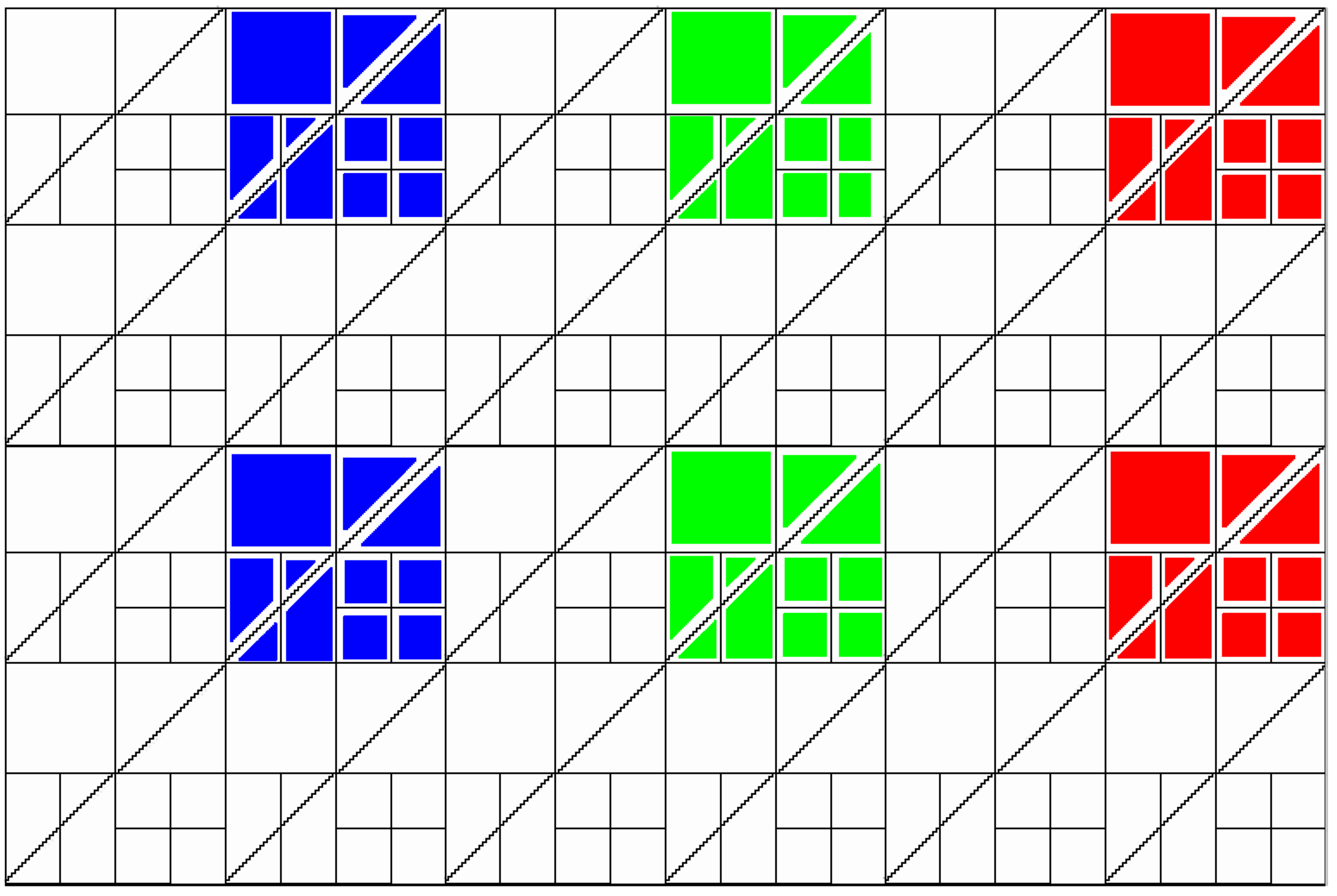} }
\subfigure[Ideal result for three classes]{\label{idealSim2}\includegraphics[angle=0, width=6cm]{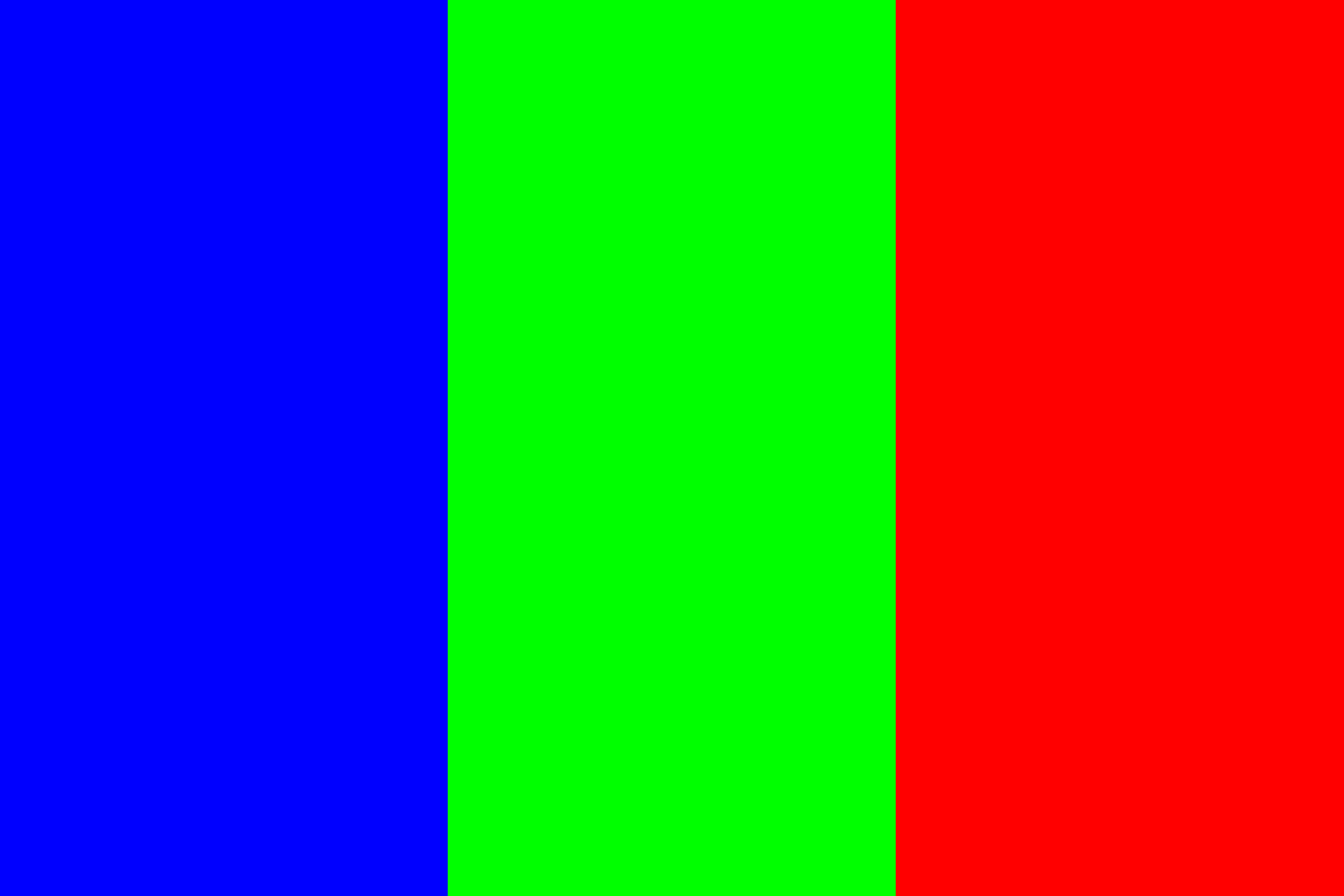} }
}
\caption{Simulated data.}\label{simCasesAndIdeal}
\end{figure}

Furthermore, although the simulated images have six well-defined blocks of targets (regions), such objects where classified in two scenarios (i.e., classes configurations).
The first scenario considers each block as a single class, the second scenario considers the blocks $1$ and $4$; $2$ and $5$; and $3$ and $6$ as three distinct classes. 
The second scenarios describes situations where the user specifies less classes than those actually present in the image.
According to this organization, Figures~\ref{samplesSim1} and~\ref{idealSim1} present the training samples and the ideal results expected for the first scenario;  similarly, Figures~\ref{samplesSim2} and~\ref{idealSim2} refer to the second scenario.

Using preliminary tests, the order of the R\'{e}nyi distance ($\beta$) was set to $0.9$ for both SVM and MSDC methods.
The SVM penalty and kernel parameter were adjusted for each image, considering a fixed parameter space, through an exhaustive search for the configuration which produces the most accurate results with respect to testing samples. 
In this study, penalty ranges in $\left\lbrace 1, 10, 100, 1000, 10000  \right\rbrace$ and the kernel flexibility $\gamma$ in $\left\lbrace 0.05, 0.1, 0.15, \ldots, 10.0 \right\rbrace$. 

Two multiclass strategies were considered for SVM in order to assess relationships between strategies, scenarios, kernels and method performance: 
One-Against-All (OAA) and One-Against-One (OAO), denoted SVM-OAA and SVM-OAO respectively.
Details about these strategies can be found in~\citet{Webb2011}.

\begin{figure}[hbt!]
\centering
\subfigure[Six classes classification]{\label{resSim6}\includegraphics[angle=0, width=13cm]{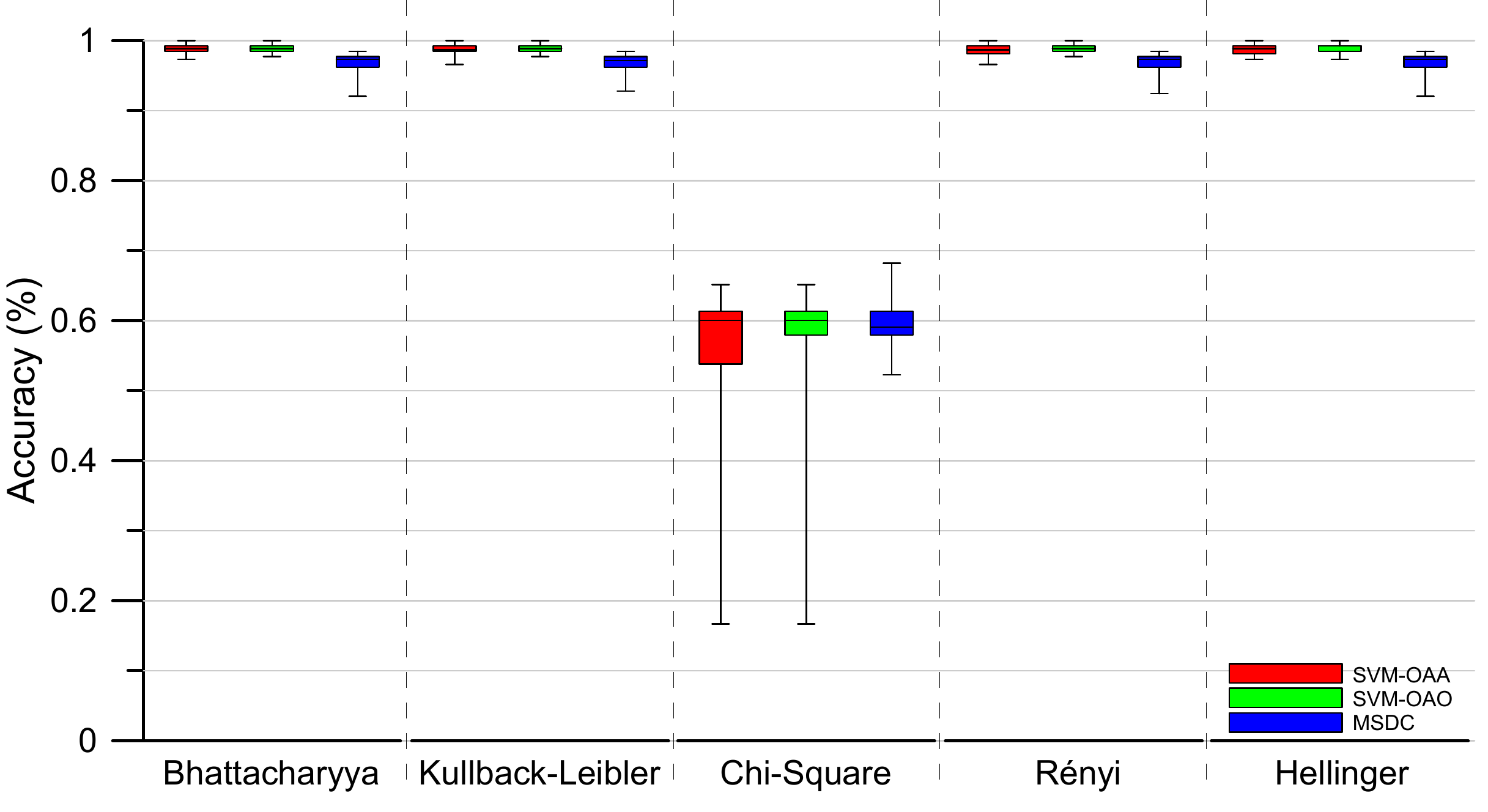} } \\
\subfigure[Three classes classification]{\label{resSim3}\includegraphics[angle=0, width=13cm]{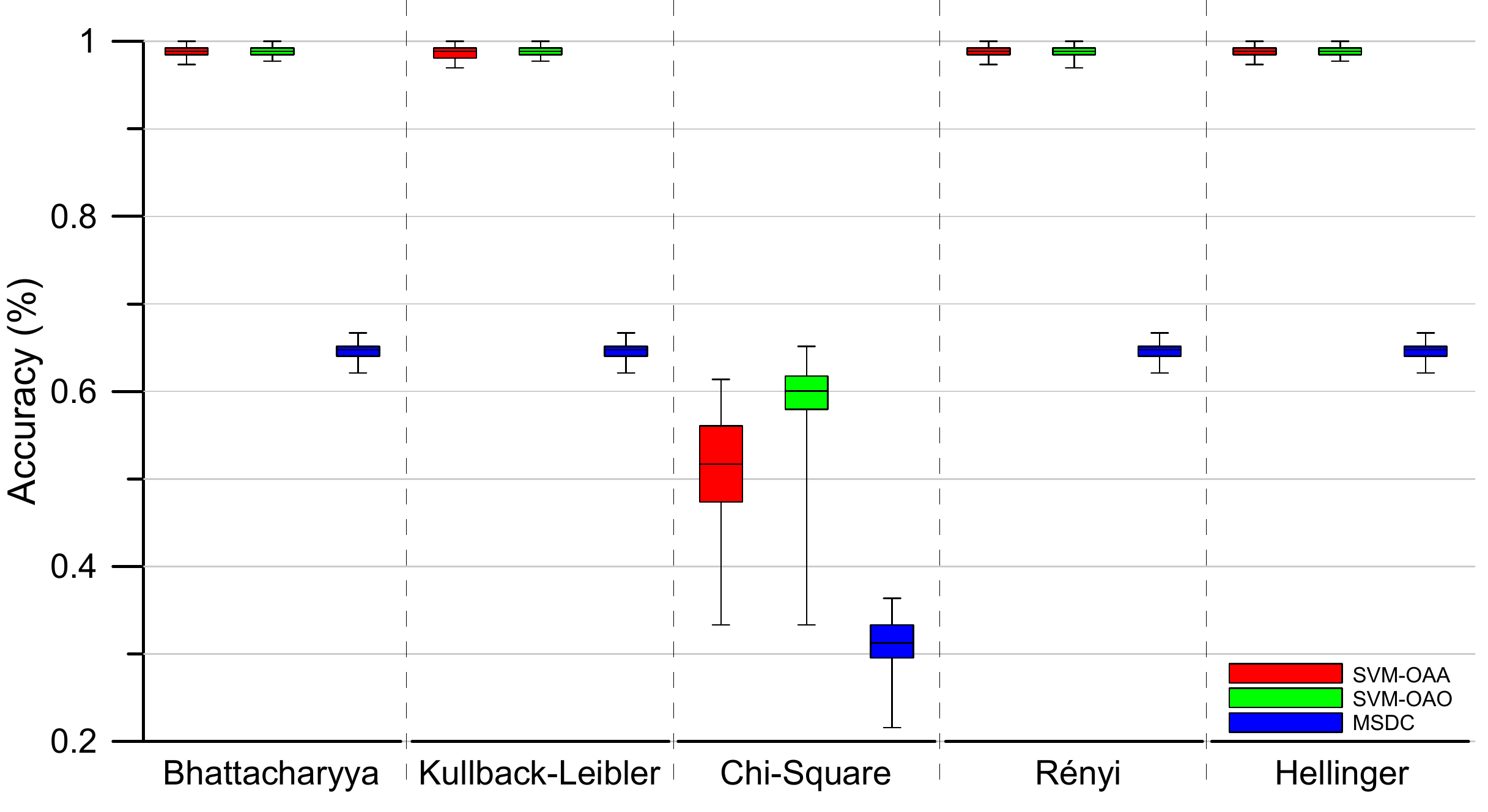} }
\caption{The accuracy of classification results for the simulated data set.}\label{resSimulados}
\end{figure}

We measured the accuracy of each classification result by the number of correctly classified regions, without taking into account training regions.
Figure~\ref{resSimulados} represents the classification accuracy achieved by each method for different configurations (i.e., stochastic distances/kernel and multiclass strategy for SVM). 
Table~\ref{tabTH_Sim} presents the $p$-values of a bilateral $t$-test to check the statistical equality between the accuracy values achieved by two distinct combinations of methods and distances. 
Further discussions about statistical equality are based on \SI{95}{\percent} of confidence. 

In the following tables SA, SO and MS represent the SVM-OAA, SVM-OAO and MSDC methods, and B, K, C, R and H represent the Bhattacharyya, Kullback-Leibler, Chi-Square, R\'{e}nyi and Hellinger stochastic distances in the kernel functions.

\begin{table}[hbt]

\caption{Test statistic $p$-values for the $t$ test that verifies that two classification techniques produce equivalent results.
Values above (below) the diagonal correspond to the six (three, resp.) classes. 
Underlined values indicate equivalent coefficients at the \SI{95}{\percent} level.}\label{tabTH_Sim}
\centering
{\tinyv

\begin{tabular}{cc|ccc|ccc|ccc|ccc|ccc|}
 &  & \multicolumn{3}{c|}{B} & \multicolumn{3}{c|}{K} & \multicolumn{3}{c|}{C} & \multicolumn{3}{c|}{R} & \multicolumn{3}{c|}{H}\tabularnewline
 &  & SA & SO & MS & SA & SO & MS & SA & SO & MS & SA & SO & MS & SA & SO & MS\tabularnewline
\thickhline
\multirow{3}{*}{B} & SA & -- & \underline{.808} & \underline{.191} & \underline{.929} & \underline{.872} & \underline{.186} & .001 & .000 & .000 & \underline{.927} & \underline{.850} & \underline{.189} & \underline{.979} & \underline{.850} & \underline{.191}\tabularnewline
 & SO & \underline{.914} & -- & \underline{.147} & \underline{.757} & \underline{.936} & \underline{.142} & .001 & .000 & .000 & \underline{.752} & \underline{.957} & \underline{.145} & \underline{.797} & \underline{.966} & \underline{.147}\tabularnewline
 & MS & .000 & .000 & -- & \underline{.228} & \underline{.162} & \underline{.984} & .002 & .000 & .000 & \underline{.224} & \underline{.156} & \underline{.987} & \underline{.203} & \underline{.161} & \underline{1.00}\tabularnewline
\hline 
\multirow{3}{*}{K} & SA & \underline{.861} & \underline{.785} & .000 & -- & \underline{.815} & \underline{.224} & .001 & .000 & .000 & \underline{1.00} & \underline{.795} & \underline{.227} & .000 & \underline{.796} & \underline{.228}\tabularnewline
 & SO & \underline{.978} & \underline{.935} & .000 & \underline{.840} & -- & \underline{.157} & .001 & .000 & .000 & \underline{.811} & \underline{.979} & \underline{.160} & \underline{.858} & \underline{.973} & \underline{.162}\tabularnewline
 & MS & .000 & .000 & \underline{.996} & .000 & .000 & -- & .002 & .000 & .000 & \underline{.220} & \underline{.151} & \underline{.997} & \underline{.199} & \underline{.156} & \underline{.984}\tabularnewline
\hline 
\multirow{3}{*}{C} & SA & .000 & .000 & \underline{.064} & .000 & .000 & \underline{.064} & -- & \underline{.881} & \underline{.691} & .001 & .001 & .002 & .001 & .001 & .002\tabularnewline
 & SO & .000 & .000 & \underline{.355} & .000 & .000 & \underline{.335} & \underline{.478} & -- & \underline{.785} & .000 & .000 & .000 & .000 & .000 & .000\tabularnewline
 & MS & .000 & .000 & .000 & .000 & .000 & .000 & .016 & .001 & -- & .000 & .000 & .000 & .000 & .000 & .000\tabularnewline
\hline 
\multirow{3}{*}{R} & SA & \underline{.903} & \underline{.822} & .000 & \underline{.955} & \underline{.881} & .000 & .000 & .000 & .000 & -- & \underline{.790} & \underline{.223} & \underline{.949} & \underline{.792} & \underline{.224}\tabularnewline
 & SO & \underline{.986} & \underline{.906} & .000 & \underline{.880} & \underline{.996} & .000 & .000 & .000 & .000 & \underline{.921} & -- & \underline{.154} & \underline{.837} & \underline{.993} & \underline{.156}\tabularnewline
 & MS & .000 & .000 & \underline{.996} & .000 & .000 & \underline{1.00} & \underline{.064} & \underline{.355} & .000 & .000 & .000 & -- & \underline{.202} & \underline{.159} & \underline{.987}\tabularnewline
\hline 
\multirow{3}{*}{H} & SA & \underline{.972} & \underline{.889} & .000 & .000 & \underline{.951} & .000 & .000 & .000 & .000 & \underline{.933} & \underline{.987} & .000 & -- & \underline{.875} & \underline{.203}\tabularnewline
 & SO & \underline{.949} & \underline{.964} & .000 & \underline{.814} & \underline{.971} & .000 & .000 & .000 & .000 & \underline{.853} & \underline{.938} & .000 & \underline{.886} & -- & \underline{.161}\tabularnewline
 & MS & .000 & .000 & \underline{1.00} & .000 & .000 & \underline{.996} & \underline{.064} & \underline{.335} & .000 & .000 & .000 & \underline{.996} & .000 & .000 & --\tabularnewline
\thickhline 
\end{tabular}
}
\end{table}

Focusing on the results of the first scenario (six classes), illustrated in Figure~\ref{resSim6}, we obtain high accuracy values regardless the method or distance adopted, except for the Chi-Square distance.
Six hundred classifications were produced by 
	three methods ({SVM-OAA}, {SVM-OAO} and {MSDC}), 
	four distances (Bhattacharyya, Kullback-Leibler, R\'{e}nyi and Hellinger, except Chi-Square) 
	and fifty synthetic images.
The minimum and maximum values observed over the \num{600} classifications were \SI{92}{\percent} and \SI{100}{\percent} respectively.

The Chi-Square distance does not only produce lower accuracy, but also higher variation in comparison to the other distances/methods. 
Numerical problems with the Chi-square distance have been reported by \citet{FreryEA2011}.
Furthermore, the use of such distance in {MSDC} and {SVM} through~\eqref{distC} provides statistically equal results.

{SVM} has better performance than {MSDC} in the second scenario. 
Usually, OAO multiclass strategy provides higher average accuracy compared to OAA, even though both strategies provide statistically equal results.
Furthermore, we note that the methods are not influenced by the adopted distance, with exception of the Chi-Square distance, where the results are very similar.
It is noteworthy that Batthacharyya, R\'{e}nyi and Hellinger distances in {SVM} lead to accurate results and small deviations.

\begin{figure}[hbt!]
\centering
\includegraphics[angle=0, width=13cm]{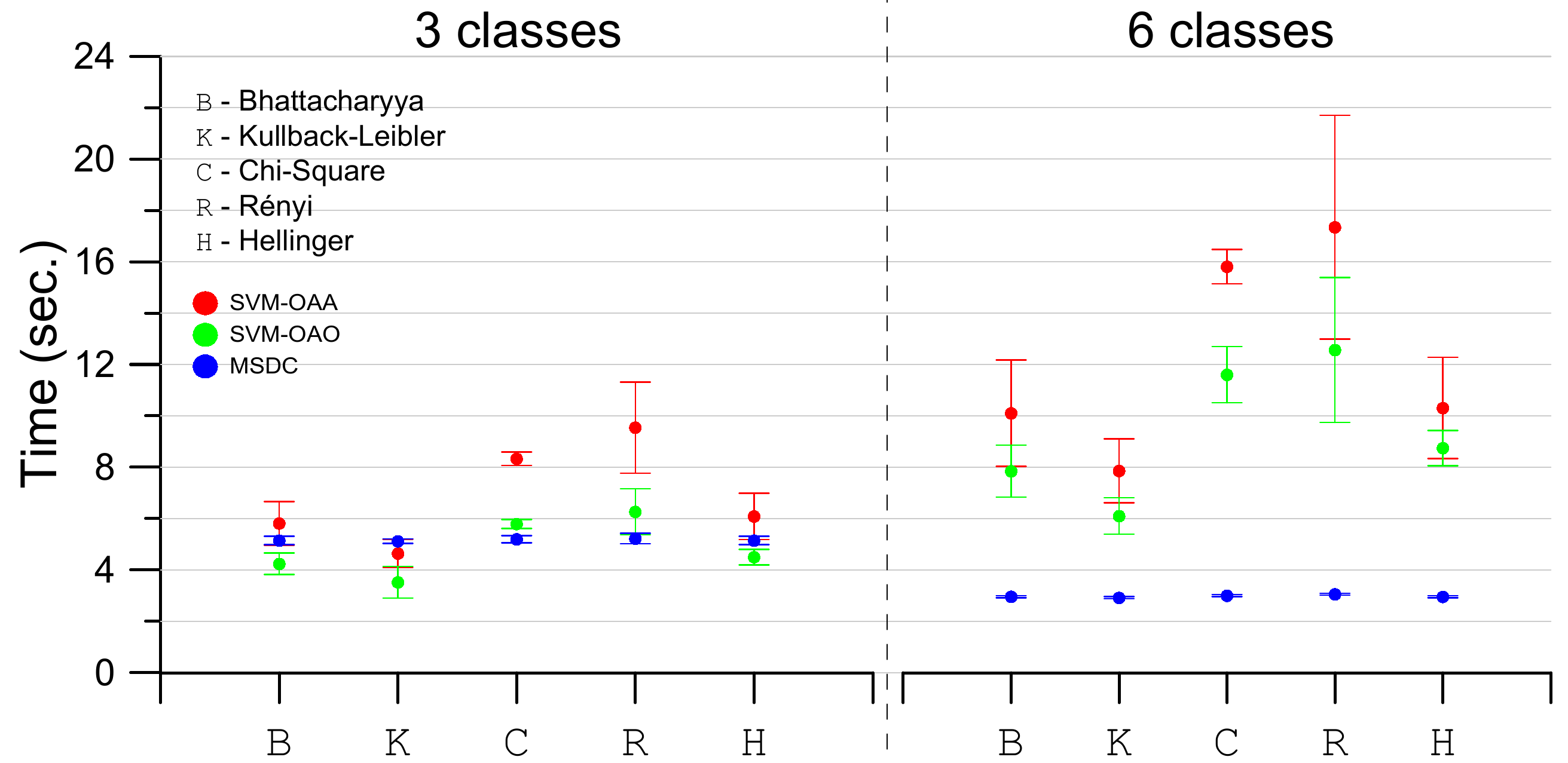}
\caption{The computational of the analyzed methods in the expriment with synthetic data.}\label{timeSim}
\end{figure}

Figure~\ref{timeSim} presents the computational time spent for training and performing the classification by each method. 
We observe that in the first scenario the {MSDC} method spends approximately \SI{3}{\second}, while {SVM} has a higher cost, specially when OAA strategy is adopted.
The reason for this is the training stage, which although SVM with OAO strategy requires splitting the multiclass classification in fifteen binary problems, when OAA is adopted the SVM training needs to solve six large quadratic optimization problems \citep[for details, see][]{TK2008}.

Additionally, we note that {MSDC} it is more expensive in the second scenario than in the first. 
Although the second scenario has fewer classes, the quantity of training regions is larger and, then, requires more time to estimate the parameters (i.e., the covariance matrix of each classes) of the probability distribution function that models the classes. 
With respect to {SVM}, in general it requires less time in comparison to the first scenario since there are less classes. 
OAA is still more expensive in the second scenario. 
Furthermore, the Kullback-Leibler kernel function requires the shortest average times {SVM}.

\subsection{Classification of Data from an Actual Sensor} \label{realcase}

This section presents classification results of the ALOS-PALSAR image shown in  Figure~\ref{ImgAE}. 
In analogy to Section~\ref{simClaRes}, we discuss a variety of classification scenarios. 

The first scenario uses all LULC classes identified in the study area; cf.\ Table~\ref{tabSamples}. 
The union of the agriculture classes (i.e., A1, A2 and A3) defines the new class called Agricultural Areas (AA).
A second scenario is created with the five classes AA, PF, PS, RG and BS. 
The last third scenario consists of three classes: Agricultural Areas, High Biomass (HB) and Low Biomass (LB). 
HB is obtained merging Primary Forest and Regeneration classes, LB comes from the union between Pasture and Bare Soil. 
We obtain the training and testing samples of AA, HB and LB classes by merging the sample polygons of the individual classes. 
These scenarios describe practical situations of users with different interests and knowledge of the area.
The Appendix provides the sample covariance matrices from the six LULC classes.

\begin{figure}[hbt]
\centering
\mbox{
\subfigure[Scenario 1]{\label{samplesS1}\includegraphics[angle=0, width=4.2cm]{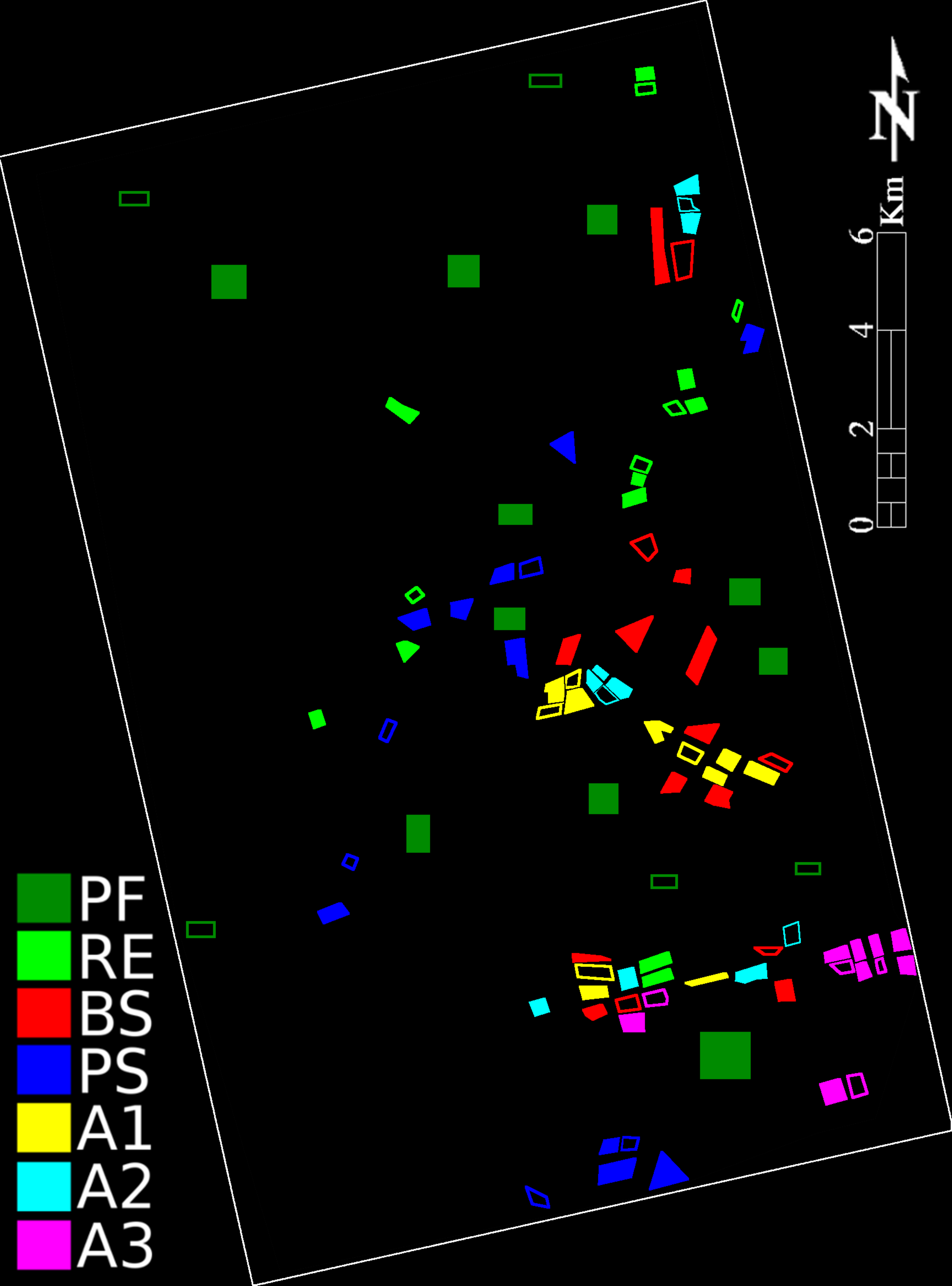} }
\subfigure[Scenario 2]{\label{samplesS2}\includegraphics[angle=0, width=4.2cm]{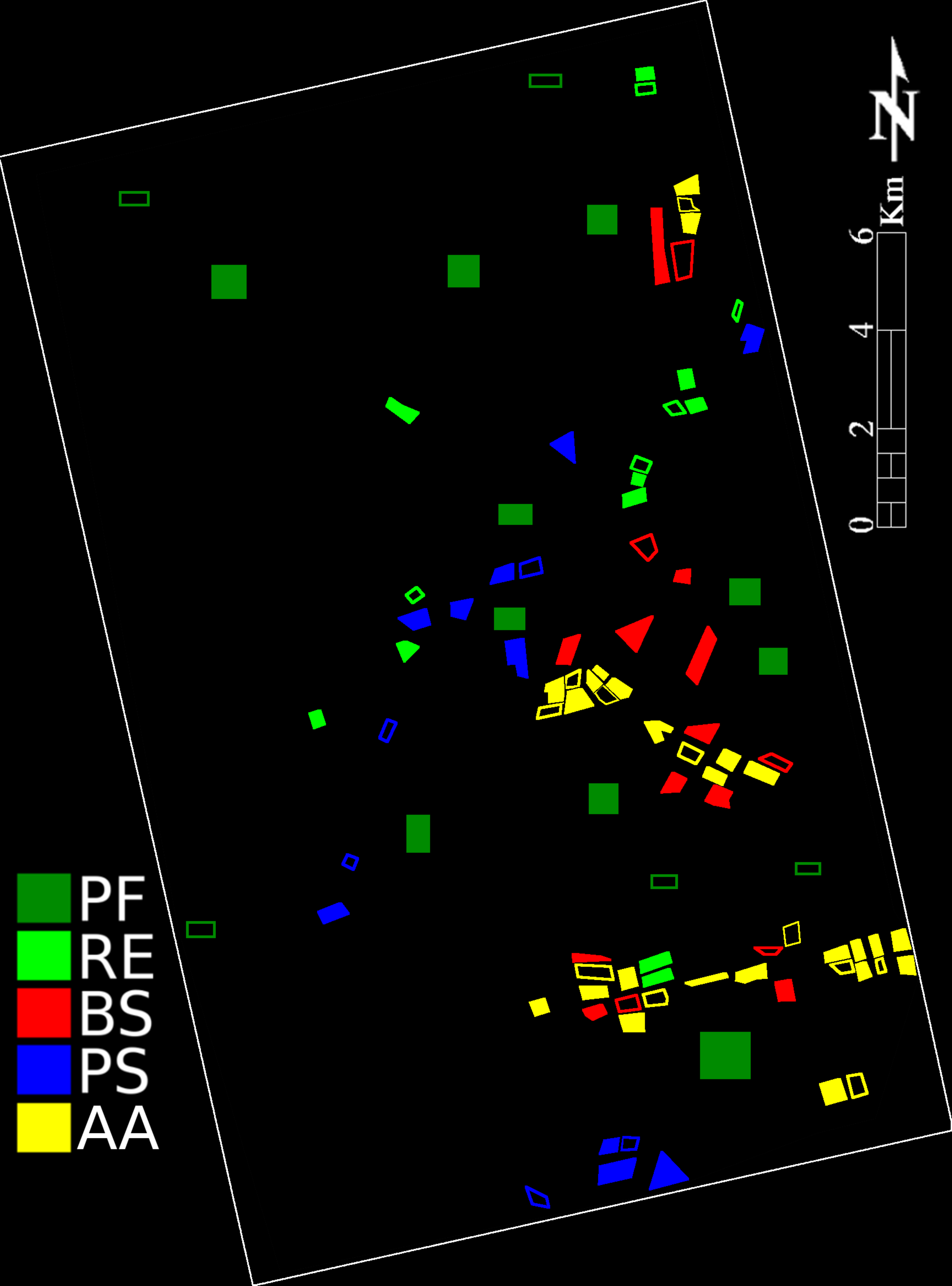} } } \\
\mbox{
\subfigure[Scenario 3]{\label{samplesS3}\includegraphics[angle=0, width=4.2cm]{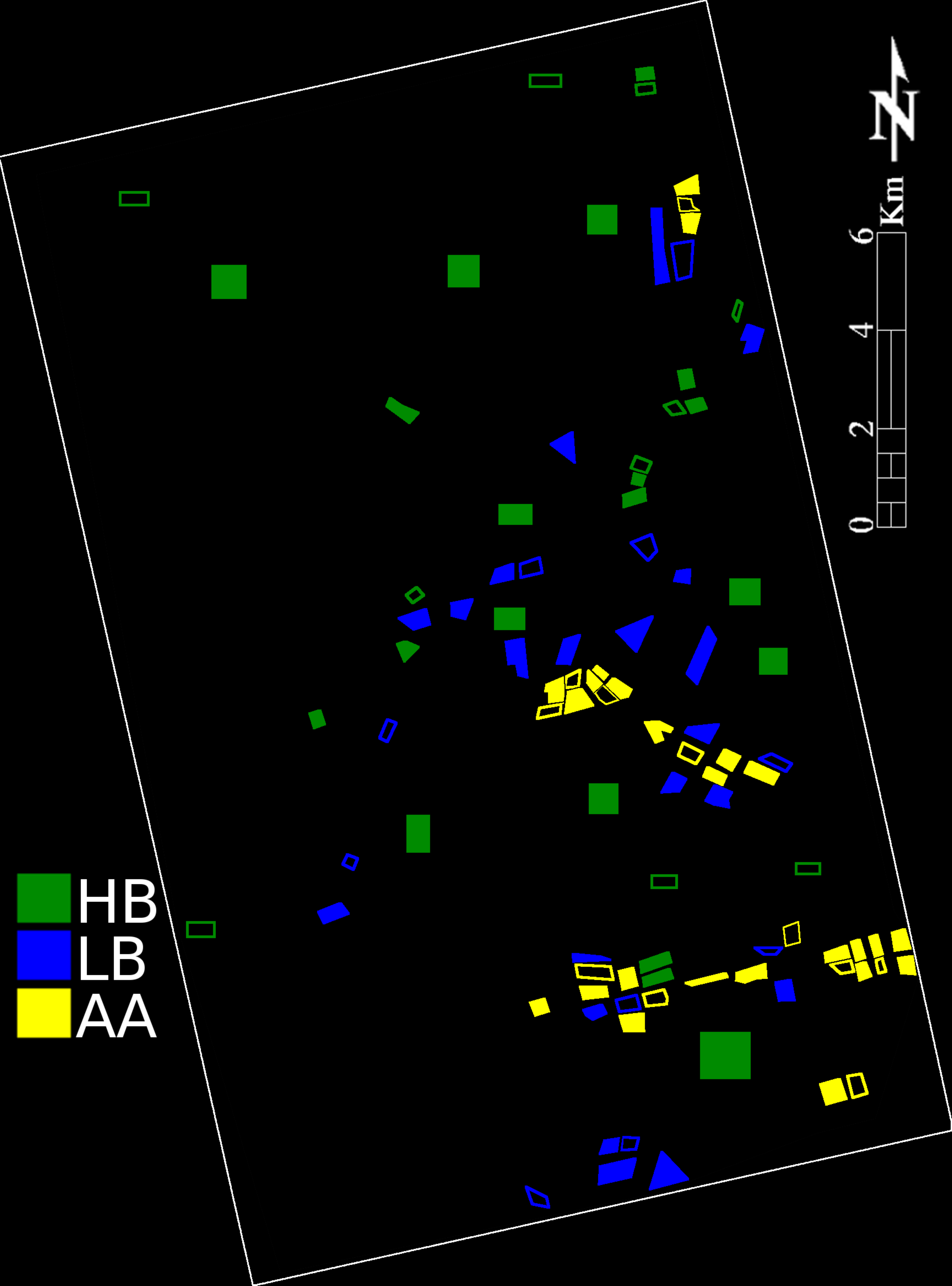} }
\subfigure[Segmentation]{\label{segmAE}\includegraphics[angle=0, width=4.2cm]{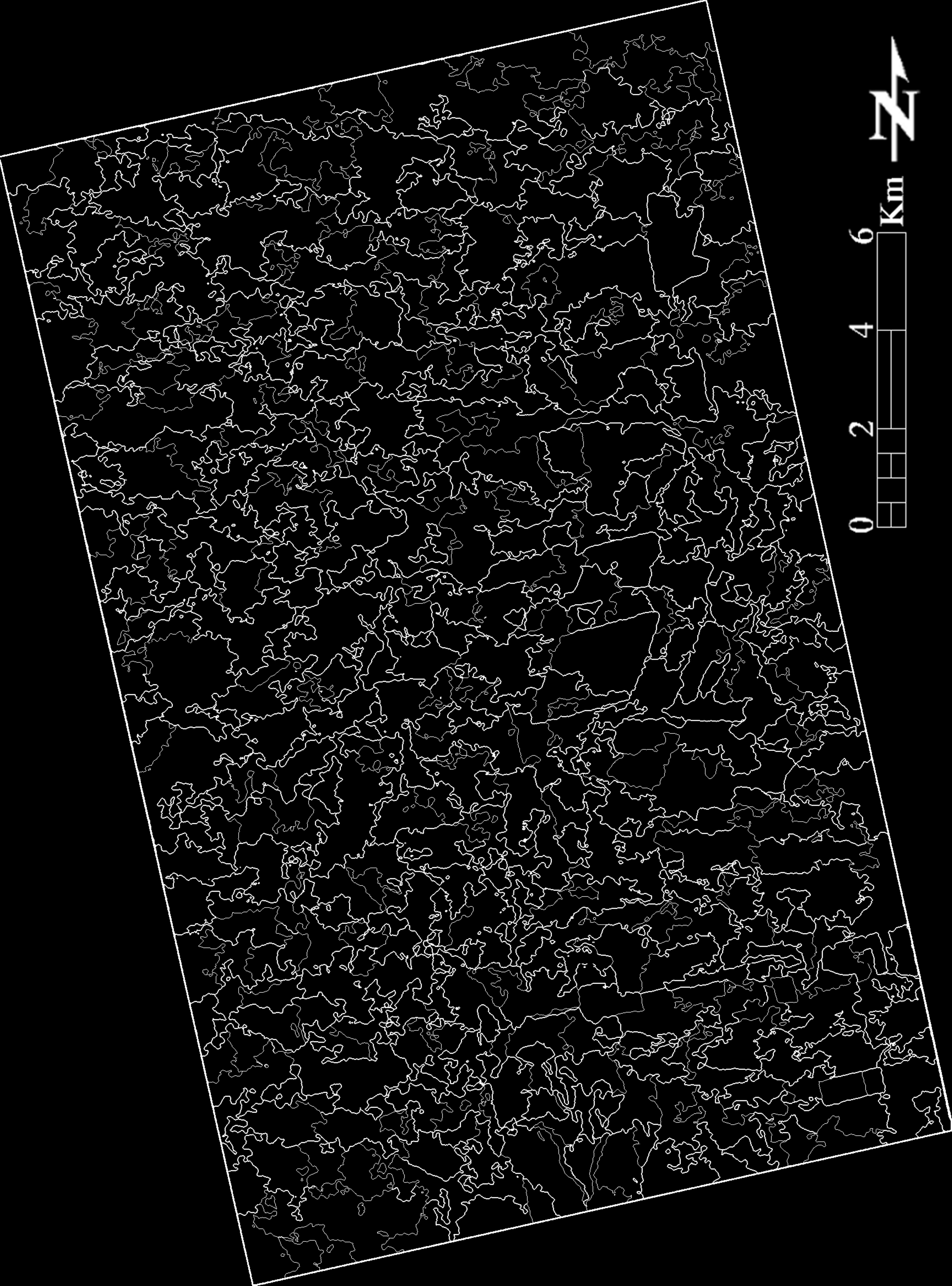} } }
\caption{Spatial distribution of the samples on the different considered scenarios and the adopted segmentation.}\label{SamplesSegm}
\end{figure}

Figures~\ref{samplesS1}, \ref{samplesS2} and \ref{samplesS3} show the spatial distribution of training and testing samples of each scenario, while Table~\ref{tabSamples} presents a summary of the LULC samples. 
The spatial distribution of these samples is also shown in Figure~\ref{samplesS1}, where training and test samples are shown in solid and empty polygons, respectively.

The region-based classification requires a segmentation.
It was performed using the region-growing method available in the Geographic Information System SPRING \citep[][freely available at \url{http://www.dpi.inpe.br/spring/english/}]{CamaraEA1996}, choosing the segmentation parameter by visual inspection. 
Figure~\ref{segmAE} shows the segments contours.

\begin{table}[hbt]
\caption{Summary of the land cover classes samples.}\label{tabSamples}
\centering
\begin{tabular}{cccrr}
\thickhline
\multirow{2}{*}{LULC Classes} & \multicolumn{2}{c}{Training} & \multicolumn{2}{c}{Testing}\tabularnewline
\cline{2-3} \cline{4-5}
 & Polygons & Pixels & Polygons & Pixels\tabularnewline
\hline 
Agruculture 1 (A1) & 4 & 3669 & 8 & 7455\tabularnewline
Agruculture 2 (A2)& 4 & 2902 & 8 & 6731\tabularnewline
Agruculture 3 (A3)& 3 & 2332 & 8 & 7049\tabularnewline
Primary Forest (PF) & 3 & 5430 & 10 & 29306\tabularnewline
Pasture (PS) & 5 & 3334 & 10 & 12866\tabularnewline
Regeneration (RE) & 5 & 2570 & 10 & 7307\tabularnewline
Bare Soil (BS) & 5 & 5384 & 11 & 13352\tabularnewline
\thickhline
\end{tabular}
\end{table}

We applied all possible combinations of methods, distances and multiclass strategies to the image and its segmentation.
The SVM parameters were obtained following the same procedures and space searches described in Section~\ref{simClaRes}.
The data were spatially subsampled taking one every three pixels in both horizontal and vertical direction in order to reduce the spatial dependence. 
The accuracy was measured by the kappa agreement coefficient with respect to the test samples.

\begin{figure}[htb]
\centering
{\includegraphics[angle=0, width=13cm]{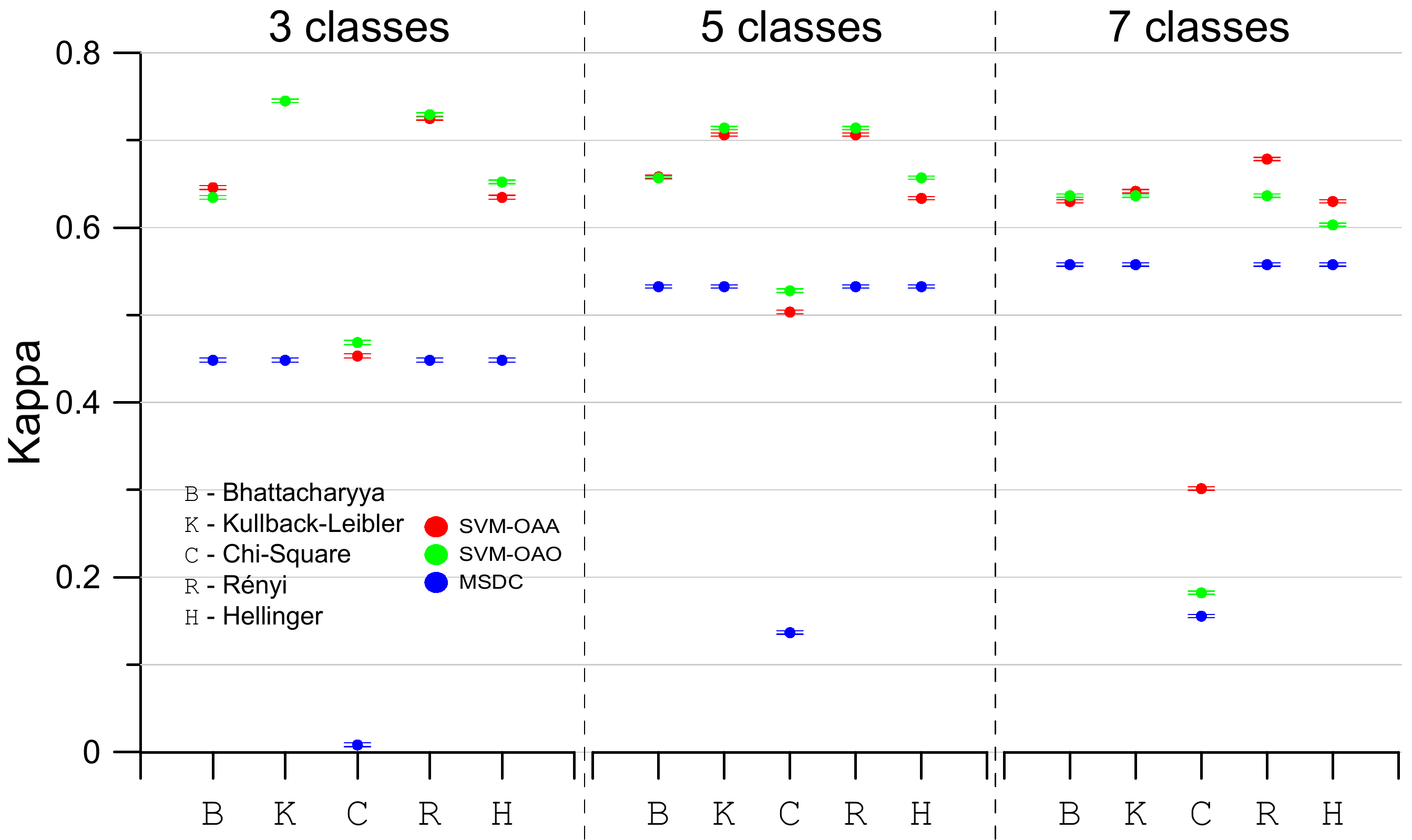} }
\caption{Classification accuracy.}\label{resReal}
\end{figure}

Figure~\ref{resReal} shows the kappa values along with their standard deviation. 
Additionally, Tables~\ref{tabTH_real7}, \ref{tabTH_real5} and \ref{tabTH_real3} present the $p$-values of a bilateral hypothesis test to check the statistical equality between kappa values achieved by two distinct combinations of methods and distances for each scenario. 
Further discussions about statistical equality are based on \SI{95}{\percent} of confidence.

\begin{table}[htb]
\caption{$p$-values from hypothesis test for comparing methods and distances with seven classes. Underlined values indicate equivalent coefficients at the \SI{95}{\percent} level.}\label{tabTH_real7}
\centering
{\tinyv
\begin{tabular}{cc|ccc|ccc|ccc|ccc|ccc|}
 &  & \multicolumn{3}{c|}{B} & \multicolumn{3}{c|}{K} & \multicolumn{3}{c|}{C} & \multicolumn{3}{c|}{R} & \multicolumn{3}{c|}{H}\tabularnewline
 &  & SA & SO & MS & SA & SO & MS & SA & SO & MS & SA & SO & MS & SA & SO & MS\tabularnewline
\thickhline
\multirow{3}{*}{B} & SA & -- & .014 & .000 & .000 & .014 & .000 & .000 & .000 & .000  & .000 & .014 & .000 & \underline{1.00} & .000 & .000\tabularnewline
 & SO &  & -- & .000 & \underline{.052} & \underline{1.00} & .000 & .000 & .000 & .000  & .000 & \underline{1.00} & .000 & .014 & .000 & .000\tabularnewline
 & MS &  &  & -- & .000 & .000 & \underline{1.00} & .000 & .000 & .000  & .000 & .000 & \underline{1.00} & .000 & .000 & \underline{1.00}\tabularnewline
\hline 
\multirow{3}{*}{K} & SA &  &  &  & -- & \underline{.052} & .000 & .000 & .000 & .000  & .000 & \underline{.052} & .000 & .000 & .000 & .000\tabularnewline
 & SO &  &  &  &  & -- & .000 & .000 & .000 & .000  & .000 & \underline{1.00} & .000 & .014 & .000 & .000\tabularnewline
 & MS &  &  &  &  &  & -- & .000 & .000 & .000  & .000 & .000 & \underline{1.00} & .000 & .000 & \underline{1.00}\tabularnewline
\hline 
\multirow{3}{*}{C} & SA &  &  &  &  &  &  & -- & .000 & .000  & .000 & .000 & .000 & .000 & .000 & .000\tabularnewline
 & SO &  &  &  &  &  &  &  & -- & .000 & .000 & .000 & .000 & .000 & .000 & .000\tabularnewline
 & MS &  &  &  &  &  &  &  &  & -- & .000 & .000 & .000 & .000 & .000 & .000\tabularnewline
\hline 
\multirow{3}{*}{R} & SA &  &  &  &  &  &  &  &  &  & -- & .000 & .000 & .000 & .000 & .000\tabularnewline
 & SO &  &  &  &  &  &  &  &  &  &  & -- & .000 & .014 & .000 & .000\tabularnewline
 & MS &  &  &  &  &  &  &  &  &  &  &  & -- & .000 & .000 & \underline{1.00}\tabularnewline
\hline 
\multirow{3}{*}{H} & SA &  &  &  &  &  &  &  &  &  &  &  &  & -- & .000 & .000\tabularnewline
 & SO &  &  &  &  &  &  &  &  &  &  &  &  &  & -- & .000\tabularnewline
\thickhline
\end{tabular}
}
\end{table}

\begin{table}[hbt]
\caption{$p$-values from hypothesis for comparing methods and distances with five classes. Underlined values indicate equivalent coefficients at the \SI{95}{\percent} level.}\label{tabTH_real5}
\centering
{\tinyv
\begin{tabular}{cc|ccc|ccc|ccc|ccc|ccc|}
 &  & \multicolumn{3}{c|}{B} & \multicolumn{3}{c|}{K} & \multicolumn{3}{c|}{C} & \multicolumn{3}{c|}{R} & \multicolumn{3}{c|}{H}\tabularnewline
 &  & SA & SO & MS & SA & SO & MS & SA & SO & MS & SA & SO & MS & SA & SO & MS\tabularnewline
\thickhline
\multirow{3}{*}{B} & SA & -- & \underline{.697} & .000 & .000 & .000 & .000 & .000 & .000 & .000 & .000 & .000 & .000 & .000 & \underline{.697} & .000\tabularnewline
 & SO &  & -- & .000 & .000 & .000 & .000 & .000 & .000 & .000 & .000 & .000 & .000 & .000 & \underline{1.00} & .000\tabularnewline
 & MS &  &  & -- & .000 & .000 & \underline{1.00} & .00 & \underline{.100} & .000 & .000 & .000 & \underline{1.00} & .000 & .000 & \underline{1.00}\tabularnewline
\hline 
\multirow{3}{*}{K} & SA &  &  &  & -- & .002 & .000 & .000 & .000 & .000 & \underline{1.00} & .002 & .000 & .000 & .000 & .000\tabularnewline
 & SO &  &  &  &  & -- & .000 & .000 & .000 & .000 & .002 & \underline{1.00} & .000 & .000 & .000 & .000\tabularnewline
 & MS &  &  &  &  &  & -- & .000 & \underline{.100} & .000 & .000 & .000 & \underline{1.00} & .000 & .000 & \underline{1.00} \tabularnewline
\hline 
\multirow{3}{*}{C} & SA &  &  &  &  &  &  & -- & .000 & .000 & .000 & .000 & .000 & .000 & .000 & .000\tabularnewline
 & SO &  &  &  &  &  &  &  & -- & .000 & .000 & .000 & \underline{.100} & .000 & .000 & \underline{.100} \tabularnewline
 & MS &  &  &  &  &  &  &  &  & -- & .000 & .000 & .000 & .000 & .000 & .000\tabularnewline
\hline 
\multirow{3}{*}{R} & SA &  &  &  &  &  &  &  &  &  & -- & .002 & .000 & .000 & .000 & .000\tabularnewline
 & SO &  &  &  &  &  &  &  &  &  &  & -- & .000 & .000 & .000 & .000\tabularnewline
 & MS &  &  &  &  &  &  &  &  &  &  &  & -- & .000 & .000 & \underline{1.00}\tabularnewline
\hline 
\multirow{3}{*}{H} & SA &  &  &  &  &  &  &  &  &  &  &  &  & -- & \underline{.500} & .000\tabularnewline
 & SO &  &  &  &  &  &  &  &  &  &  &  &  &  & -- & .000\tabularnewline
\thickhline
\end{tabular}
}
\end{table}

\begin{table}[hbt]
\caption{$p$-values from hypothesis tests for comparing methods and distances with three classes. Underlined values indicate equivalent coefficients at the \SI{95}{\percent} level.}\label{tabTH_real3}
\centering
{\tinyv
\begin{tabular}{cc|ccc|ccc|ccc|ccc|ccc|}
 &  & \multicolumn{3}{c|}{B} & \multicolumn{3}{c|}{K} & \multicolumn{3}{c|}{C} & \multicolumn{3}{c|}{R} & \multicolumn{3}{c|}{H}\tabularnewline
 &  & SA & SO & MS & SA & SO & MS & SA & SO & MS & SA & SO & MS & SA & SO & MS\tabularnewline
\thickhline
\multirow{3}{*}{B} & SA & -- & .000 & .000 & .000 & .000 & .000 & .000 & .000 & .000 & .000 & .000 & .000 & .000 & .031 & .000\tabularnewline
 & SO &  & -- & .000 & .000 & .000 & .000 & .000 & .000 & .000 & .000 & .000 & .000 & \underline{.970} & .000 & .000\tabularnewline
 & MS &  &  & -- & .000 & .000 & \underline{1.00} & \underline{.154} & .000 & .000 & .000 & .000 & \underline{1.00} & .000 & .000 & \underline{1.00} \tabularnewline
\hline 
\multirow{3}{*}{K} & SA &  &  &  & -- & \underline{.069} & .000 & .000 & .000 & .000 & .000 & .000 & .000 & .000 & .000 & .000\tabularnewline
 & SO &  &  &  &  & -- & .000 & .000 & .000 & .000 & .000 & .000 & .000 & .000 & .000 & .000\tabularnewline
 & MS &  &  &  &  &  & -- & \underline{.154} & .000 & .000 & .000 & .000 & \underline{1.00} & .000 & .000 & \underline{1.00} \tabularnewline
\hline 
\multirow{3}{*}{C} & SA &  &  &  &  &  &  & -- & .000 & .000 & .000 & .000 & \underline{.154} & .000 & .000 & \underline{.154} \tabularnewline
 & SO &  &  &  &  &  &  &  & -- & .000 & .000 & .000 & .000 & .000 & .000 & .000\tabularnewline
 & MS &  &  &  &  &  &  &  &  & -- & .000 & .000 & .000 & .000 & .000 & .000\tabularnewline
\hline 
\multirow{3}{*}{R} & SA &  &  &  &  &  &  &  &  &  & -- & \underline{.105} & .000 & .000 & .000 & .000\tabularnewline
 & SO &  &  &  &  &  &  &  &  &  &  & -- & .000 & .000 & .000 & .000\tabularnewline
 & MS &  &  &  &  &  &  &  &  &  &  &  & -- & .000 & .000 & \underline{1.00}\tabularnewline
\hline 
\multirow{3}{*}{H} & SA &  &  &  &  &  &  &  &  &  &  &  &  & -- & .000 & .000\tabularnewline
 & SO &  &  &  &  &  &  &  &  &  &  &  &  &  & -- & .000\tabularnewline
\thickhline
\end{tabular}
}
\end{table}

We observe that the Chi-Square distance produces low kappa values in both methods and with both multiclass strategies for {SVM}. 
This is due to the aforementioned numerical instabilities presented by this measure. 
While most of the considered distances ranged, in the experiments, from $10^{-1}$ to $10^2$, the Chi-Square had its values approximately in $0$ to $10^6$.

Results provided by {MSDC} using the Bhatacharyya, Kullback-Leibler, R\'{e}nyi and Hellinger are statistically the same. 

Similarly to the results presented in Section~\ref{simClaRes}, the choice of a multiclass strategy does not have strong influence on the performance of SVM.
Except when the Batthacharyya distance is used, the increased in intra-class variability, which occurs when the number of classes decrease, suggests the use of OAO strategy in SVM.
Observing the SVM performance as function of the stochastic distance integrated in its kernel, the R\'{e}nyi distance has the highest accuracy in the first scenario.
Regarding the second and third scenarios, SVM performs better when the kernel functions are enhanced with Kullback-Leibler and R\'{e}nyi distances.

The influence of the scenario is noteworthy.
As the number of classes decreases, leading to increasing intra-class variability, the performance of MSDC also decreases. 
Converseley, the classification accuracy of SVM tends to increase while the number of classes decrease.

In summary, {SVM} presented better performance with respect to {MSDC}. 
Kullback-Leibler and R\'{e}nyi distances are the preferred choice for defining radial basis kernel functions for the region-based classification with {SVM}. 

\begin{figure}[hbt!]
\centering
\mbox{
\subfigure[SA/B -- Scenario 1]{\label{s1-OAA}\includegraphics[angle=0, width=4.2cm]{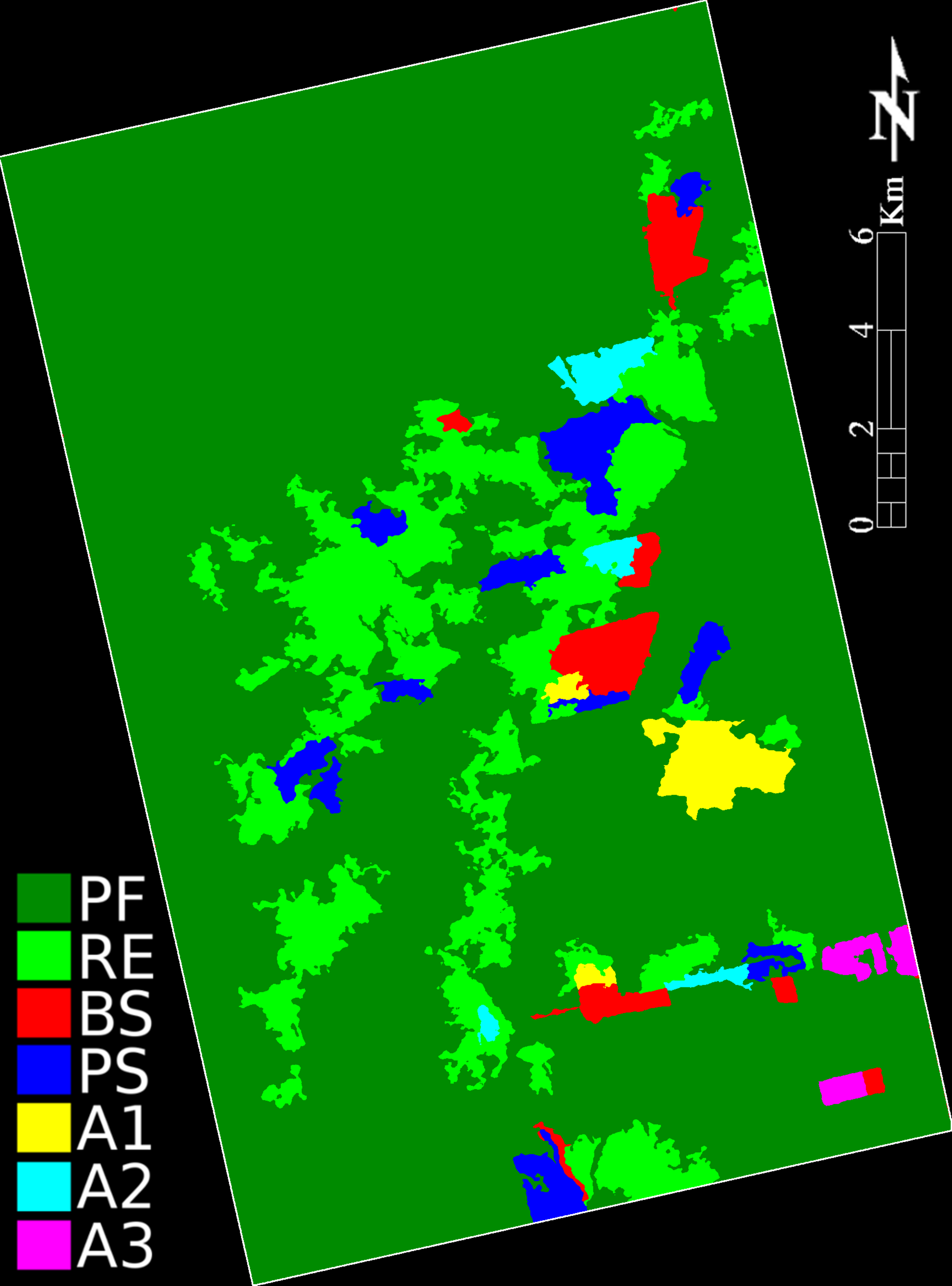} }
\subfigure[SO/K -- Scenario 1]{\label{s1-OAO}\includegraphics[angle=0, width=4.2cm]{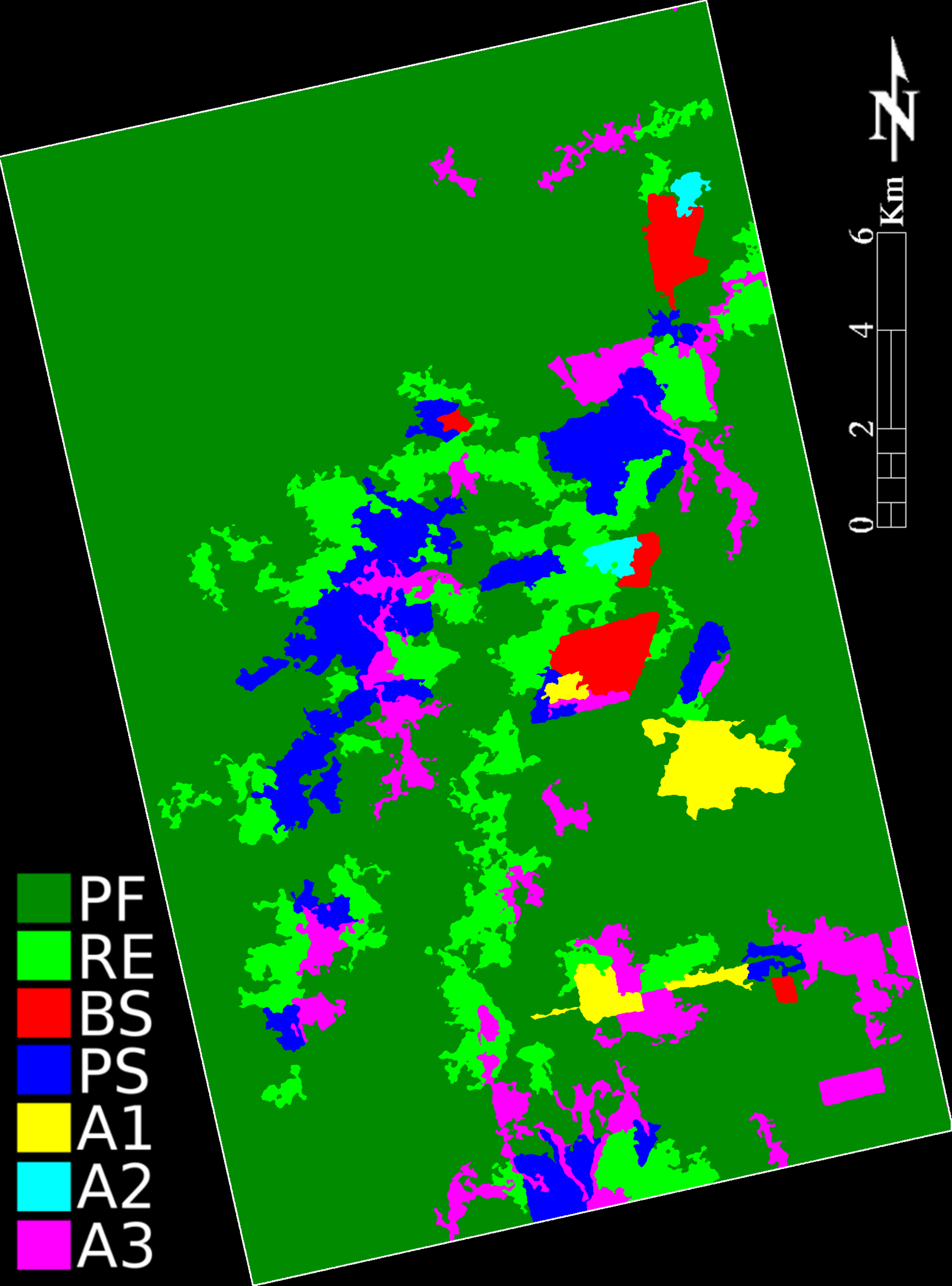} }
\subfigure[MS/R -- Scenario 1]{\label{s1-MSDC}\includegraphics[angle=0, width=4.2cm]{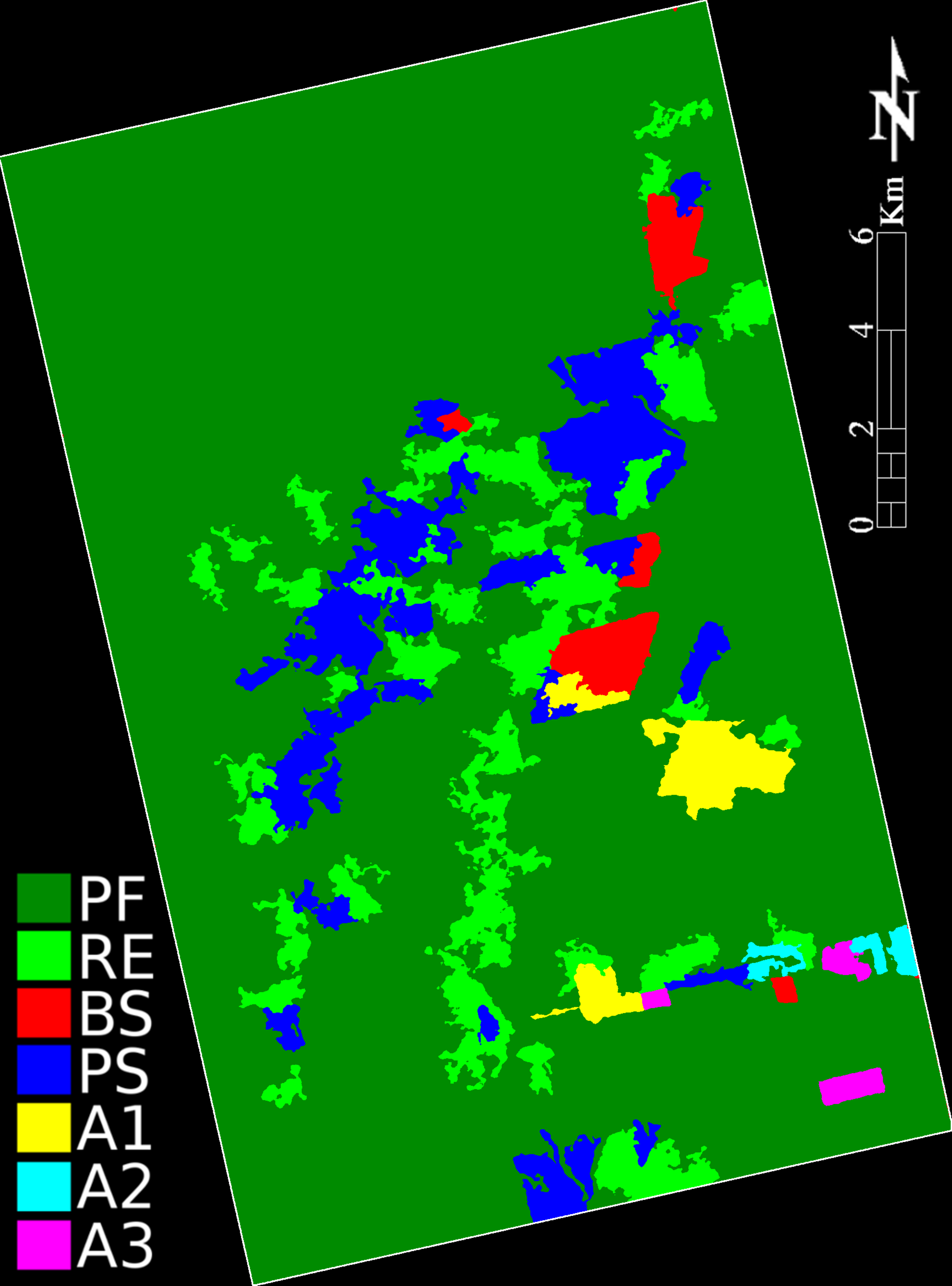} }
} \\

\mbox{
\subfigure[SA/H -- Scenario 2]{\label{s2-OAA}\includegraphics[angle=0, width=4.2cm]{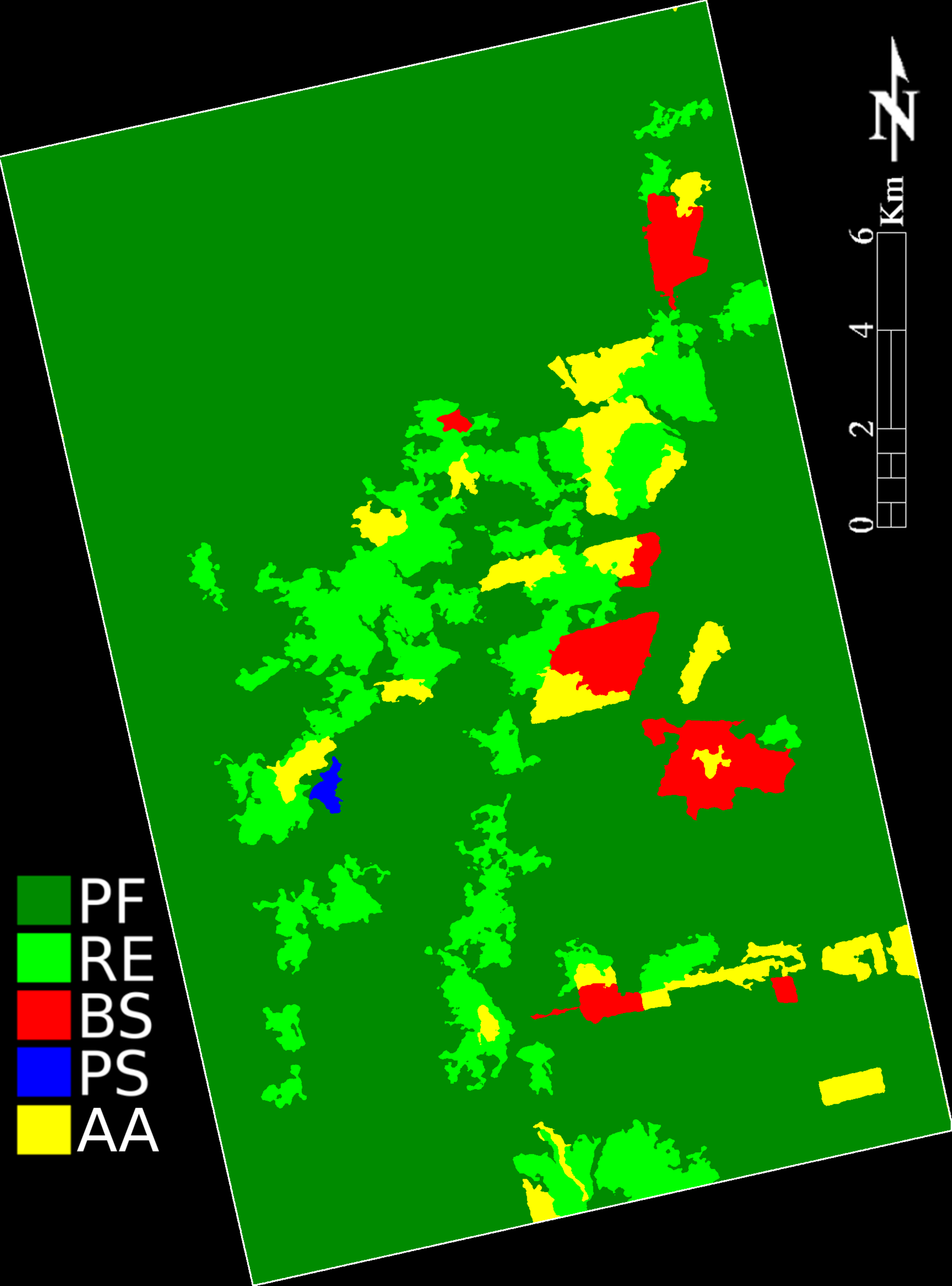} }
\subfigure[SO/R -- Scenario 2]{\label{s2-OAO}\includegraphics[angle=0, width=4.2cm]{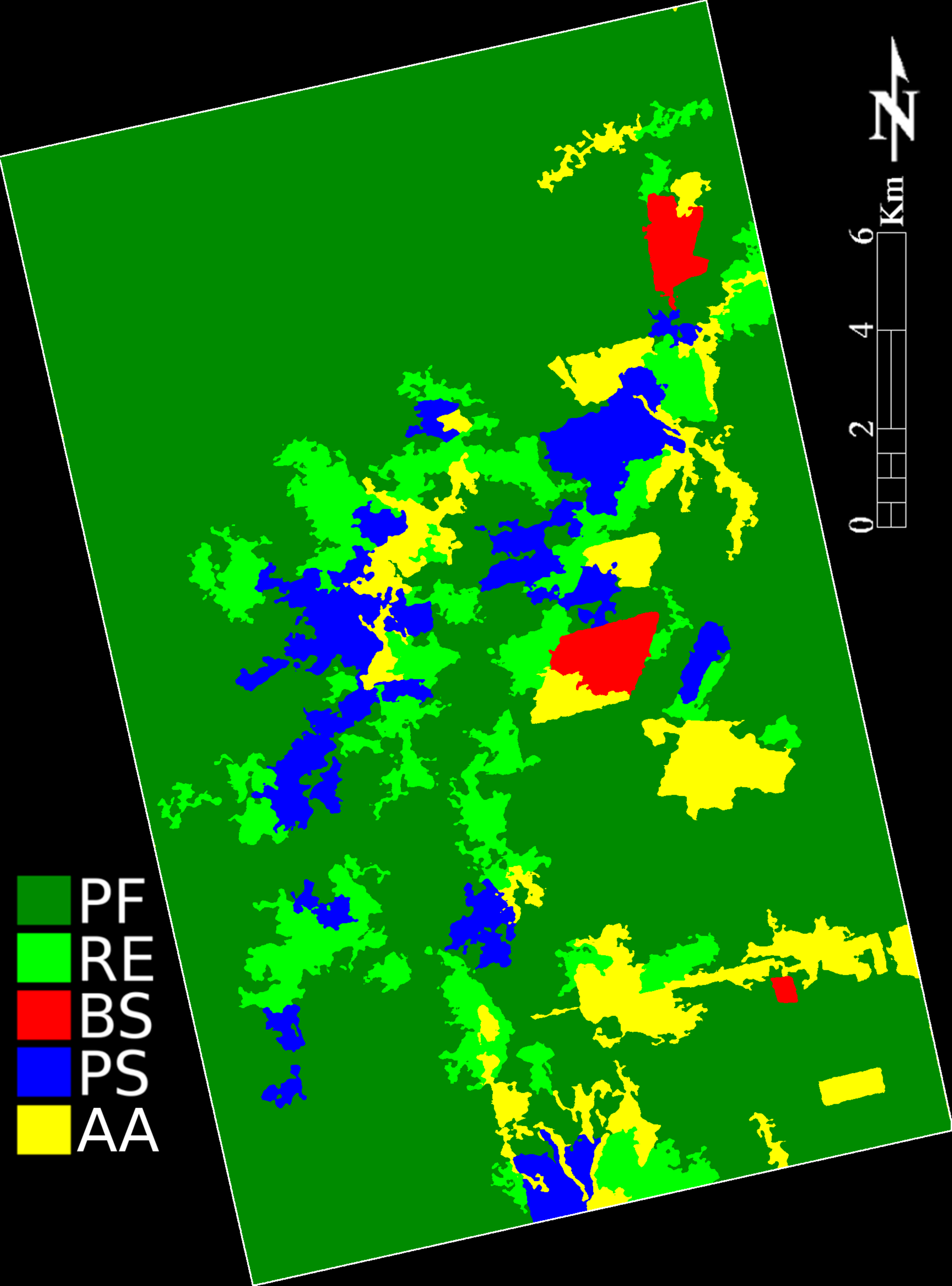} }
\subfigure[MS/K -- Scenario 2]{\label{s2-MSDC}\includegraphics[angle=0, width=4.2cm]{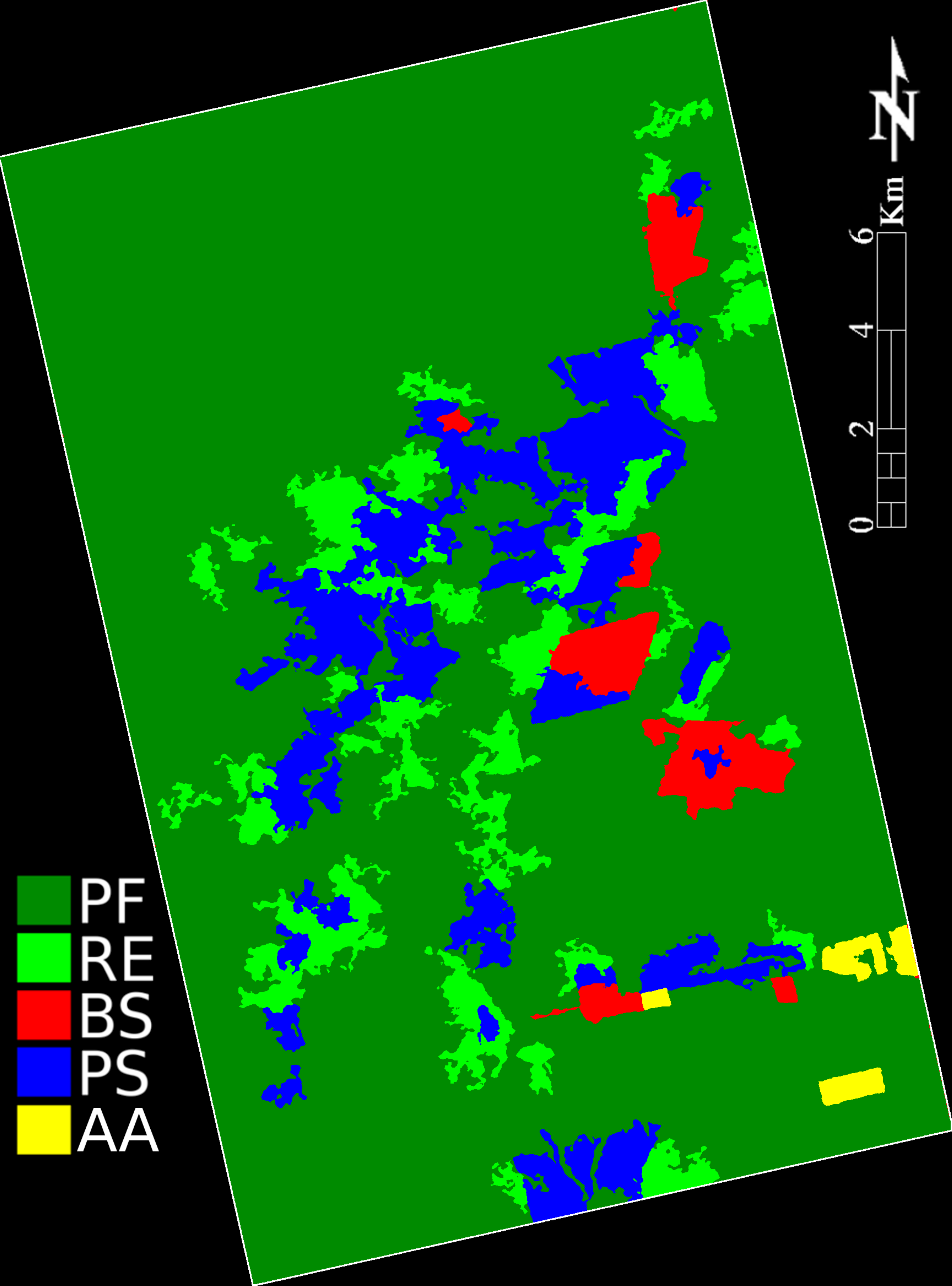} }
} \\

\mbox{
\subfigure[SA/R -- Scenario 3]{\label{s3-OAA}\includegraphics[angle=0, width=4.2cm]{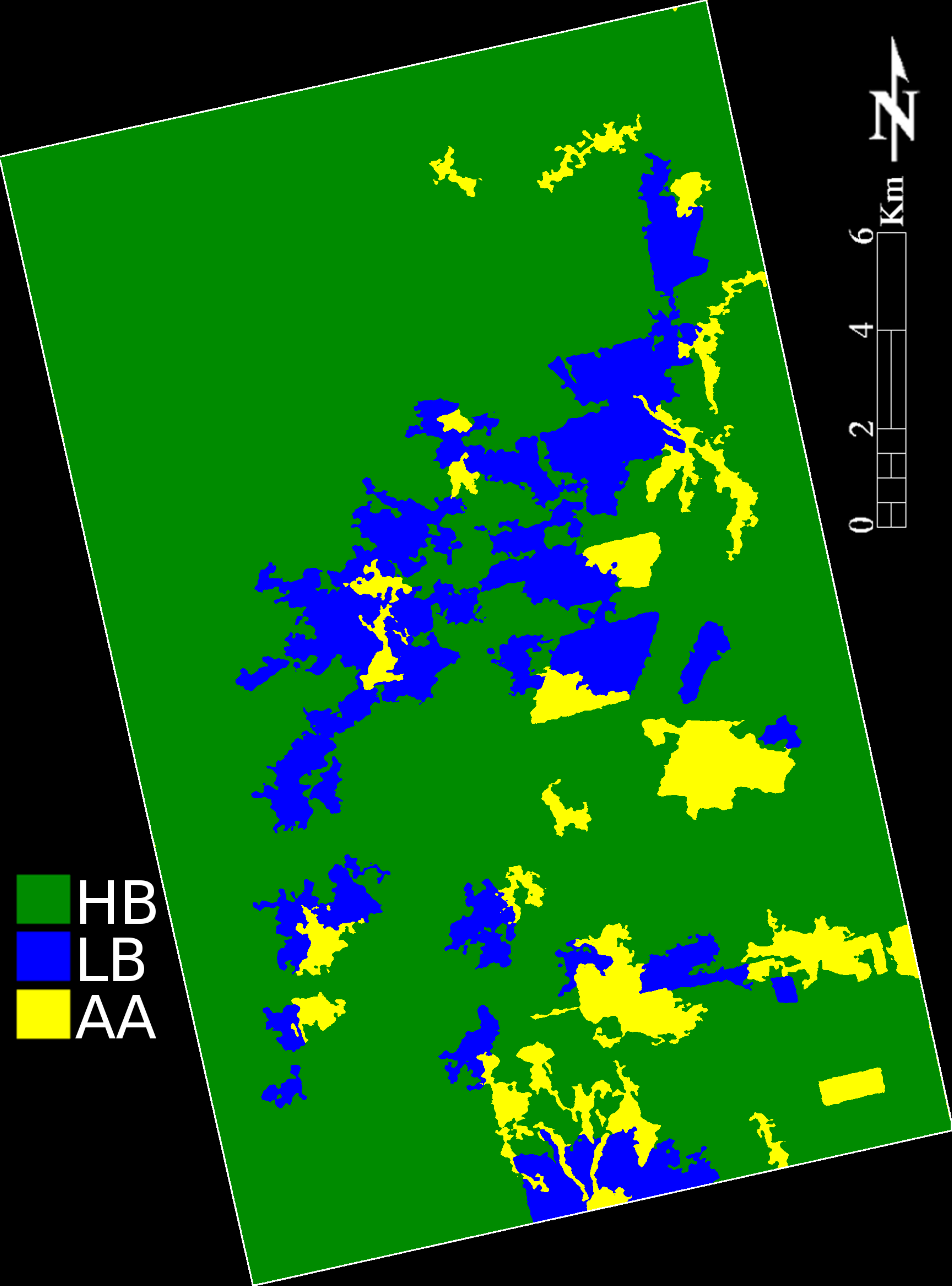} }
\subfigure[SO/K -- Scenario 3]{\label{s3-OAO}\includegraphics[angle=0, width=4.2cm]{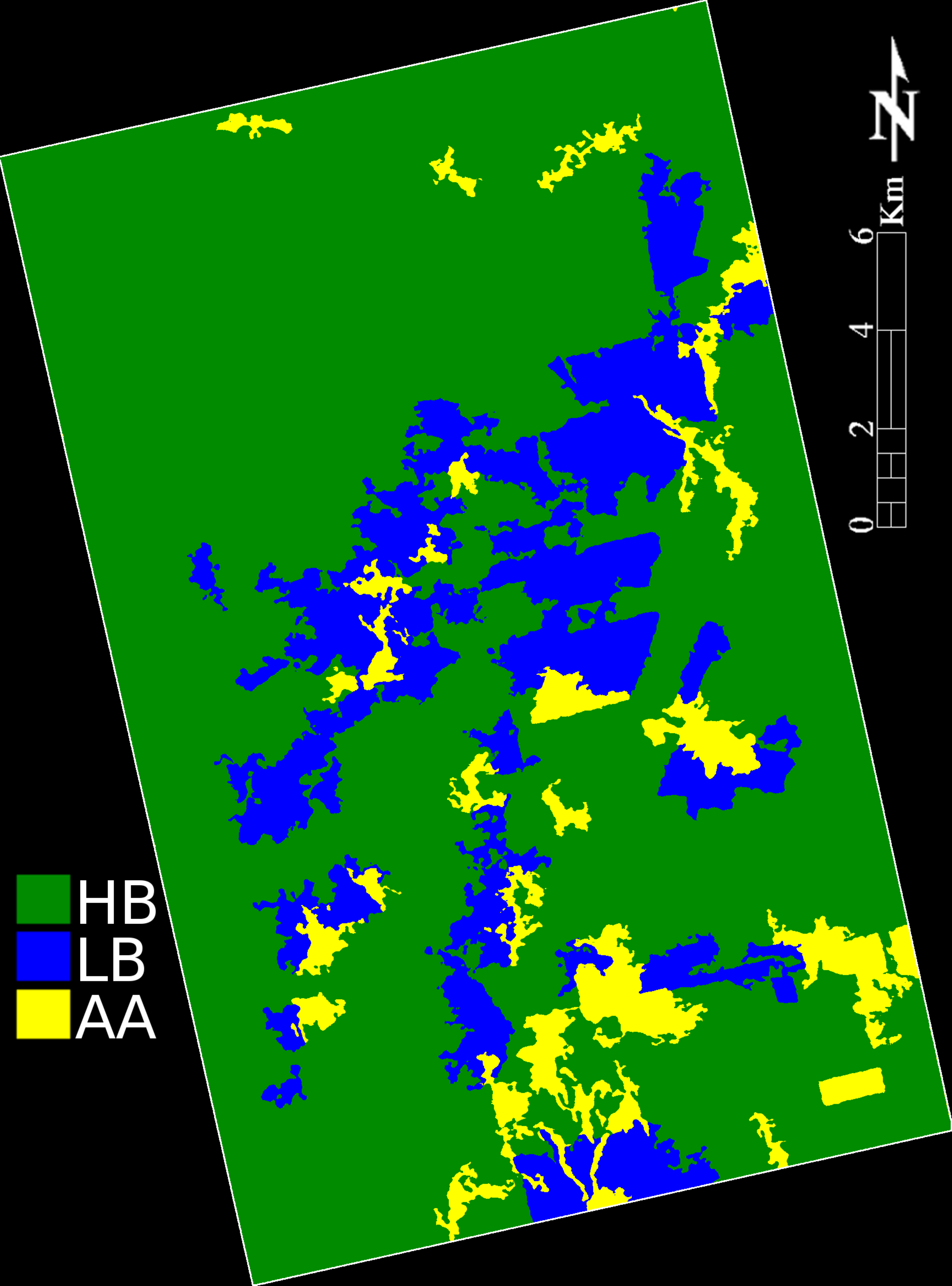} }
\subfigure[MS/H -- Scenario 3]{\label{s3-MSDC}\includegraphics[angle=0, width=4.2cm]{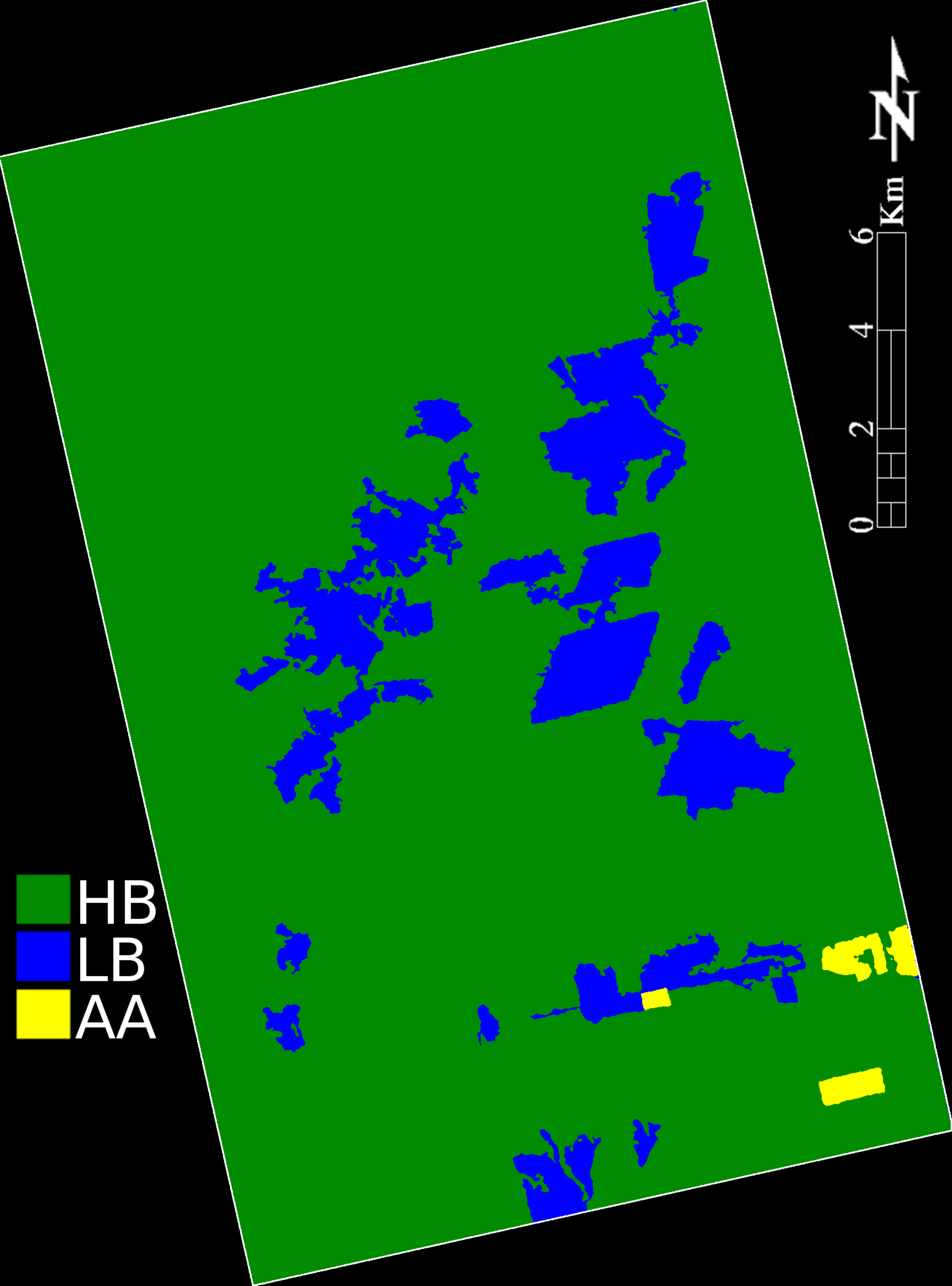} }
}
\caption{Actual data classification results.}\label{resClassReal}
\end{figure}

Figure~\ref{resClassReal} shows selected results.
The first scenario, that considers seven classes, posses a difficult problem to both methods to discriminate the agricultural classes (i.e., A1, A2 and A3). 
With respect to the second scenario, while SVM with OAA strategy does not distinguish PS and MSDC often confuses AA with PS, the SVM with OAO strategy provides a better separation between such classes. 
In the last scenario MSDC was not apt to separating AA and LH classes, differently fom SVM.
This last method, specially when using OAA strategy, provides a better classification of HB and LB areas; cf. the central region of the study area.

\begin{figure}[hbt!]
\centering
\includegraphics[angle=0, width=13cm]{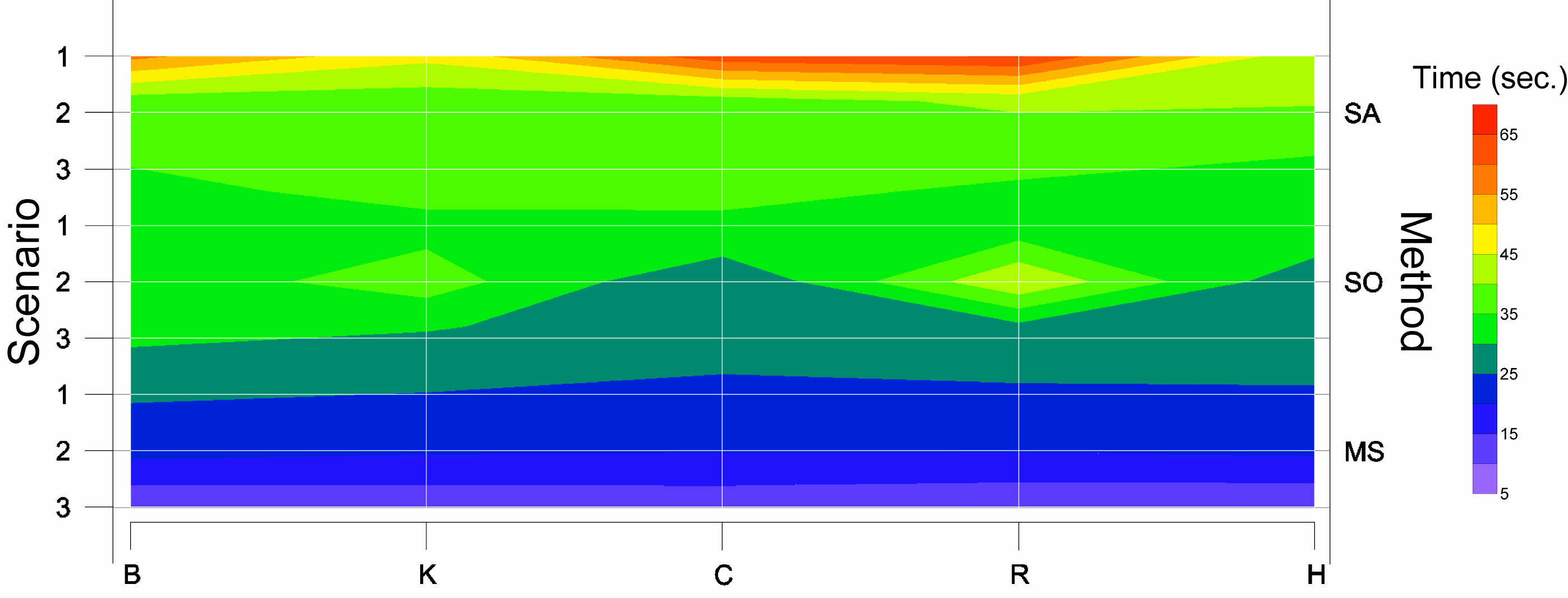}
\caption{Computational time of the analyzed methods in the experiment with actual data.}\label{timeReal}
\end{figure}

Figure~\ref{timeReal} presents the computational time spent by the methods in the experiments with actual data. 
It can be noted a gradual increase in the processing time when dealing with scenarios with more classes. 
As previously observed in the first case study, the use of OAA strategy by SVM implies more processing time compared to OAO. 
MSDC is the least computational intensive method. 
Furthermore, the time execution is relatively insensitive to choices of distances.

\section{Conclusions}\label{concl}

The objective of this study was to verify the performance of {SVM} for region-based classification of PolSAR image in comparison to {MSDC}. 
For this purpose, we adopted radial basis functions derived from stochastic distances between Complex Multivariate Wishart distributions: Bhatacharyya, Kullback-Leibler, Chi-Square, R\'{e}nyi and Hellinger. 
We used simulated images and data from an operational sensor, and proposed a number of scenarios which describe different situations of ability to discriminate classes.
These scenarios depict actual situations users encounter in practice.

It was found that {SVM} it is more robust than {MSDC} in both simulated and actual data sets because, depending on the adopted multiclass strategy, {SVM} has equal or superior performance on simple scenarios (first scenarios of simulated and actual data sets -- Figures~\ref{resSim6} and~\ref{resReal}) and superior in more complex scenarios (second scenario of simulated and second and third scenarios with actual data sets -- Figures~\ref{resSim3} and~\ref{resReal}).

The main drawbacks of {SVM} are its computational cost and the need to tuning the penalty and the kernel parameter.

As previously verified by \citet{SilvaEA2013}, the Chi-Square distance is not indicated to perform classification through {MSDC}. 
Its numerical instabilities lead to relative poor performance.

Face to the exposed results, the {SVM} method presented a better performance compared to {MSDC}. 
With respect to radial basis kernel functions considered in this study for region-based classification with {SVM}, the preferred ones are Kullback-Leibler and R\'{e}nyi distances. 
The use of distinct stochastic distances on {MSDC} does not lead to improved accuracy, so the choice should be based on computational execution time.

\section*{Disclosure statement}
There is no conflict of interest involving this research.

\section*{Funding}
The authors thank FAPESP (Grant 2014/14830-8), UNESP/PROPe (Grant 2016/1389), CNPq and Fapeal for funding this research.

\bibliographystyle{tfcad}
\bibliography{../BibTex_TESE}

\appendix
\section{Suplementary data}

The covariance matrices estimated by maximum likelihood over the LULC samples of A1, A2, A3, PF, PS, RE and BS, depictured in Figure~\ref{samplesAE}, are presented in \eqref{CovMat:A1} to \eqref{CovMat:BS}, respectively. Only the upper triangle and the diagonal are shown. The remaining elements are the complex conjugates of transposed corresponding element.


\begin{equation}\label{CovMat:A1}
\Sigma_{\textrm{A1}} = \left(
\begin{array}{ccc}
47.95 & -0.03 -0.47\textsf{i} & 7.04 +4.09\textsf{i} \\
 & 2.96 & -0.11 -0.25\textsf{i} \\
 &  & 17.39 \\
\end{array}
\right).
\end{equation}


\begin{equation}\label{CovMat:A3}
\Sigma_{\textrm{A3}} = \left(
\begin{array}{ccc}
534.48 &  2.12 +5.54\textsf{i} & 41.10 +79.48\textsf{i} \\
 & 4.59 & -1.38 +0.95\textsf{i} \\
 &  & 262.25 \\
\end{array}
\right).
\end{equation}

\begin{equation}\label{CovMat:PF}
\Sigma_{\textrm{PF}} = \left(
\begin{array}{ccc}
68.86 &  -0.32 -0.03\textsf{i} & 20.39 +1.75\textsf{i} \\
 & 20.87 & -0.49 -0.23\textsf{i} \\
 &  & 61.03 \\
\end{array}
\right).
\end{equation}

\begin{equation}\label{CovMat:PS}
\Sigma_{\textrm{PS}} = \left(
\begin{array}{ccc}
49.71 &  0.24 -0.28\textsf{i} & 22.91 -3.01\textsf{i} \\
 & 6.45 & -0.36 +0.03\textsf{i} \\
 &  & 38.50 \\
\end{array}
\right).
\end{equation}

\begin{equation}\label{CovMat:RG}
\Sigma_{\textrm{RG}} = \left(
\begin{array}{ccc}
 55.20 &  0.24 +0.15\textsf{i} & 18.51 +0.61\textsf{i} \\
  &  9.17 & -0.38 -0.14\textsf{i} \\
 &  & 35.13 \\
\end{array}
\right).
\end{equation}

\begin{equation}\label{CovMat:BS}
\Sigma_{\textrm{BS}} = \left(
\begin{array}{ccc}
21.15 &  0.01 -0.06\textsf{i} & 9.01 -1.98\textsf{i} \\
 &  2.27 & -0.03 -0.08\textsf{i} \\
 &  & 15.70 \\
\end{array}
\right).
\end{equation}

\end{document}